\documentclass[letterpaper]{article} % DO NOT CHANGE THIS
\usepackage{aaai2026}  % DO NOT CHANGE THIS
\usepackage{times}  % DO NOT CHANGE THIS
\usepackage{helvet}  % DO NOT CHANGE THIS
\usepackage{courier}  % DO NOT CHANGE THIS
\usepackage[hyphens]{url}  % DO NOT CHANGE THIS
\usepackage{graphicx} % DO NOT CHANGE THIS
\usepackage{natbib}  % DO NOT CHANGE THIS AND DO NOT ADD ANY OPTIONS TO IT
\usepackage{caption} % DO NOT CHANGE THIS AND DO NOT ADD ANY OPTIONS TO IT
\usepackage{algorithm}
\usepackage{algorithmic}
\usepackage{xcolor}
\usepackage{multirow}
\usepackage{comment}
\usepackage{amsmath}
\usepackage{amsfonts}
\usepackage{booktabs} 
\usepackage{graphicx}
\usepackage{pgfplots}
\usepackage{times}  % DO NOT CHANGE THIS
\usepackage{helvet}  % DO NOT CHANGE THIS
\usepackage{courier}  % DO NOT CHANGE THIS
\usepackage[hyphens]{url}  % DO NOT CHANGE THIS
\usepackage{graphicx} % DO NOT CHANGE THIS
\usepackage{natbib}  % DO NOT CHANGE THIS AND DO NOT ADD ANY OPTIONS TO IT
\usepackage{caption} % DO NOT CHANGE THIS AND DO NOT ADD ANY OPTIONS TO IT
\usepackage{algorithm}
\usepackage{algorithmic}
\usepackage{xcolor}
\usepackage{multirow}
\usepackage{comment}
\usepackage{pdfpages}
\usepackage{amsmath}
\usepackage{amsfonts}
\usepackage{booktabs}
\usepackage{graphicx}
\usepackage{pgfplots}
\usepackage{tabularx} % 在导言区
\usepackage{booktabs}
\usepackage{listings} % 用于自动换行的代码
\usepackage{newfloat}
\usepackage{listings}
\usepackage{bibentry}
\pgfplotsset{compat=1.18}
\usepackage{newfloat}
\usepackage{listings}

\DeclareCaptionStyle{ruled}{labelfont=normalfont,labelsep=colon,strut=off} % DO NOT CHANGE THIS
\floatstyle{ruled}
\newfloat{listing}{tb}{lst}{}
\floatname{listing}{Listing}
\nocopyright 
\usepackage{fontenc}
\title{RISE: Enhancing VLM Image Annotation with Self-Supervised Reasoning}
\author{
    Suhang Hu, Wei Hu\footnotemark[1], Yuhang Su, Fan Zhang
}
\affiliations{
    College of Information Science and Technology\\
    Beijing University of Chemical Technology
}

\usepackage{bibentry}
\newcommand{\think}[1]{\textless think\textgreater #1\textless/think\textgreater}
\newcommand{\answer}[1]{\textless answer\textgreater #1\textless/answer\textgreater}

\begin{document}

\maketitle

\footnotetext[1]{Corresponding author: Wei Hu (huwei@mail.buct.edu.cn)\\
Preprint. Under review.}

\begin{abstract}
Vision-Language Models (VLMs) struggle with complex image annotation tasks, such as emotion classification and context-driven object detection, which demand sophisticated reasoning. Standard Supervised Fine-Tuning (SFT) focuses solely on annotation outcomes, ignoring underlying rationales, while Visual Reinforcement Fine-Tuning (Visual-RFT) produces inconsistent Chains of Thought (CoTs) due to the absence of high-quality, verified CoTs during pre-training. We introduce \textbf{RISE} (Reason-Inspire-Strengthen-Expertise), a two-stage framework to overcome these limitations. In the \textbf{Reason} stage (RISE-CoT), a reinforcement learning-driven "\textbf{annotation-reasoning-annotation}" closed-loop generates visually grounded, logically consistent CoTs by verifying their ability to reconstruct original annotations without direct leakage. The \textbf{Inspire} and \textbf{Strengthen} stage (RISE-R1) leverages a high-quality CoT subset, filtered by RISE-CoT rewards, for supervised fine-tuning, followed by reinforcement fine-tuning to produce interpretable reasoning and accurate annotations, achieving \textbf{Expertise} in complex visual tasks. Evaluated on complex and simple image annotation tasks, RISE-trained Qwen2-VL-2B outperforms SFT and Visual-RFT, achieving robust performance and enhanced explainability. RISE offers a self-supervised solution for advancing VLM reasoning without requiring manually annotated CoTs. Code and resources are available at: https://github.com/HSH55/RISE.
\end{abstract}


\section{Introduction}
\textbf{Image annotation}, the process of assigning descriptive labels or semantic information to image elements, includes tasks like classification, detection, and segmentation. Traditional vision models, such as convolutional neural networks (CNNs) like ResNet~\cite{he2016deep} and YOLO~\cite{redmon2016you}, and early Transformer-based models like Vision Transformer (ViT)~\cite{dosovitskiy2020image}, excel at capturing color and texture patterns but struggle with tasks requiring complex reasoning, such as emotion classification in Emotion6~\cite{peng2015mixed} (e.g., discerning ``\textit{joy}'' from smiling faces and festive contexts) or context-driven object detection in LISA~\cite{lai2024lisa} (e.g., identifying ``\textit{a wheel not in contact with the ground}''). Vision-Language Models (VLMs), such as GPT-4v~\cite{openai2023gpt4v}, Qwen-VL~\cite{bai2023qwen}, LLaVA~\cite{liu2023visual}, and Gemini~\cite{google2023gemini}, integrate visual perception with linguistic reasoning, offering superior solutions for such tasks. However, general-purpose VLMs, especially lower-capacity models, often require task-specific fine-tuning on relevant datasets to achieve robust performance.

To enhance VLMs for complex image annotation, Supervised Fine-Tuning (\textbf{SFT}) is commonly used to predict annotations. However, SFT struggles with complex tasks, as it predicts outcomes without capturing underlying reasoning. The advent of \textbf{Chains of Thought (CoT)}—textual narratives articulating step-by-step reasoning—has shown immense potential~\cite{wei2022chain,mme-cot2024,deepseek2023deepseekr1} to improve VLM reasoning performance. CoTs enable models to produce "think-answer" outputs, enhancing both accuracy and transparency. A CoT \textit{''The scene shows smiling faces, bright colors, and festive settings}'' leads to the annotation ''\textit{joy}''. Visual Reinforcement Fine-Tuning (\textbf{Visual-RFT})~\cite{liu2025visual} and VLM-R1\cite{shen2025vlm-r1}, inspired by Open-R1 and DeepSeek-R1~\cite{deepseek2023deepseekr1}, uses reinforcement learning to produce CoTs, improving over SFT. Yet, as most datasets lack high-quality CoTs, Visual-RFT and VLM-R1 miss critical CoT pre-training, akin to DeepSeek-Zero~\cite{deepseek2023deepseekr1}, and may generate superficial CoTs (e.g., ``\textit{This is a joyful scene}'' $\rightarrow$ ``\textit{joy: 1.0}''). Our experiments and MME-CoT~\cite{mme-cot2024} confirm that low-quality CoTs reduce accuracy and generalization in complex tasks.

High-quality CoTs are essential for initializing (warming up) VLMs to excel in reasoning tasks, as demonstrated by DeepSeek-R1~\cite{deepseek2023deepseekr1}. However, generating and validating such CoTs for image annotation datasets is challenging, as manual annotation is labor-intensive and large VLMs might produce inconsistent outputs. We hypothesize that \textbf{A high-quality CoT should intrinsically encapsulate all necessary visual and contextual cues to derive the correct annotation}, such as ``The scene shows smiling faces, bright colors, and festive settings`` indicates ``joy`` for Emotion6. To address these challenges, we propose \textbf{RISE} (Reason-Inspire-Strengthen-Expertise), a \textbf{two-stage framework} that self-supervises CoT generation and trains VLMs to produce interpretable ``think-answer'' outputs, enhancing reasoning and annotation accuracy for complex tasks.

The \textbf{Reason} stage (\textbf{RISE-CoT}) is the cornerstone, employing a \textbf{reinforcement learning-driven closed-loop "annotation-reasoning-annotation"} process. Here, a VLM generates a CoT from an image and its ground-truth annotation of a training sample, then reconstructs the annotation from the generated CoT. The CoT is quantified as a score by the accuracy of this reconstruction, providing a robust self-supervised signal for optimization via Group Relative Policy Optimization (GRPO)~\cite{deepseek2024deepseekmath}. This process ensures visually grounded and logically consistent CoTs. Crucially, the CoT generation is designed to prevent direct leakage of annotations, forcing the VLM to genuinely describe the reasoning. The output of RISE-CoT is an enriched dataset, where each training sample is augmented with a CoT and its corresponding quality score, forming a comprehensive resource for subsequent fine-tuning.

The \textbf{Inspire} and \textbf{Strengthen} stage (\textbf{RISE-R1}) leverages this enriched dataset to optimize a VLM through two sequential steps. First, SFT is performed on a high-quality CoT subset, filtered based on RISE-CoT's quality scores, to initialize the VLM. Second, Reinforcement Fine-Tuning (RFT) is applied on the entire dataset, further optimizing the VLM to produce precise annotations alongside interpretable CoTs. \textbf{This VLM is deployed for image annotation tasks}.

Implemented with Qwen2-VL-2B, RISE demonstrably outperforms standard SFT and Visual-RFT in image annotation tasks. Our experiments highlight RISE’s ability to enhance VLMs’ reasoning by generating and leveraging high-quality CoTs in a self-supervised manner, ultimately underscoring the critical importance of verifiable, high-quality reasoning in multimodal learning.

\section{Related Works}

\textbf{Traditional deep learning models}, such as CNNs~\cite{lecun1989backpropagation}, YOLO~\cite{redmon2016you} for object detection, and UNet~\cite{ronneberger2015u} for segmentation, excel at pattern-based image analysis but lack natural language expression and complex reasoning capabilities. This limits the adaptability of traditional models to tasks requiring contextual understanding, such as emotion classification in Emotion6~\cite{peng2015mixed} or context-driven detection in LISA~\cite{lai2024lisa}. Our RISE framework integrates visual and linguistic reasoning to produce interpretable outputs for such complex tasks.

\textbf{Vision-Language Models (VLMs)} have transformed image understanding by combining visual perception with linguistic reasoning. CLIP~\cite{radford2021learning} achieves robust image-text alignment via contrastive learning, enabling zero-shot capabilities. Advanced models like GPT-4V~\cite{openai2023gpt4v}, Qwen-VL~\cite{bai2023qwen, qwenteam2024qwen2vl}, LLaVA~\cite{liu2023visual, liu2024llavaonevision}, and Gemini~\cite{google2023gemini, google2024gemini15} excel in tasks like Visual Question Answering and image captioning. However, VLMs often struggle with complex image annotation tasks, such as object detection~\cite{lin2014microsoft}, requiring extensive fine-tuning with task-specific data. RISE enhances small VLMs by self-supervising high-quality reasoning generation, reducing reliance on annotated datasets.

\textbf{Reasoning-enhanced models} introduce logical depth to AI systems. Chain of Thought (CoT) prompting~\cite{wei2022chain} enables step-by-step reasoning, advanced by DeepSeek-R1~\cite{deepseek2023deepseekr1} through reinforcement learning for structured ``think-answer'' outputs. Visual-RFT~\cite{liu2025visual} and VLM-R1~\cite{shen2025vlm-r1} apply this to visual tasks, using reinforcement fine-tuning to generate reasoning-augmented responses that outperform standard SFT in few-shot detection. However, their reliance on external or unverified CoTs often leads to ungrounded outputs. Approaches like LLaVA-ToT~\cite{yao2023tree} and ReAct~\cite{yao2022react} face similar issues, lacking intrinsic CoT validation. RISE overcomes this by self-supervising CoT generation and validation, ensuring visually grounded and logically consistent reasoning.

\textbf{Self-supervised learning (SSL)} mitigates reliance on annotated data. Methods like SimCLR~\cite{chen2020simple}, MoCo~\cite{he2020momentum}, MAE~\cite{he2022masked}, and DINO~\cite{caron2021emerging} learn robust visual features, while ALBEF~\cite{li2021align}, iBOT~\cite{zhou2022ibot}, and Data2Vec~\cite{baevski2022data2vec} extend SSL to multimodal settings. However, applying SSL to self-supervise complex reasoning in VLMs remains underexplored.

RISE addresses the limitations of traditional models (no reasoning), VLMs (fine-tuning needs), and reasoning-enhanced models (unverified CoTs) by self-supervising high-quality CoT generation, achieving robust performance in complex image annotation tasks.

% 主节开头：RISE和Visual-RFT适用于文字描述的image annotation任务，聚焦分类/检测
\section{RISE Framework}
\label{sec:rise}

RISE is a versatile framework designed for image annotation tasks. We assume two key prerequisites for the image annotation tasks it addresses: (1) Annotations must be \textbf{expressible as textual outputs} by the VLM and capable of being justified through textual CoTs; and (2) \textbf{Annotation correctness must be quantifiable}, allowing the computation of rewards for reinforcement learning and loss functions for supervised learning. In this paper, we demonstrate RISE’s capabilities through \textbf{two representative task types}: (1) \textbf{Classification tasks}, with annotations as probability distributions over categories; and (2) \textbf{Detection tasks}, focusing on localizing pre-specified objects with bounding boxes. While RISE is primarily evaluated on these two tasks, it is potentially applicable to simple segmentation tasks where regions are few and boundaries can be concisely described textually (e.g., as polygon contours), though such applications are left for future exploration.

RISE operates through two stages. The \textbf{RISE-CoT} stage leverages reinforcement learning in an iterative process to generate visually grounded CoTs by mapping image-annotation pairs to reasoning narratives and verifying their quality through annotation reconstruction. Subsequently, the \textbf{RISE-R1} stage builds upon these CoTs, employing SFT on a selected high-quality subset to instill robust reasoning capabilities, followed by reinforcement fine-tuning (RFT) to optimize task-specific outputs. 

\subsection{RISE-CoT: Closed-Loop Reasoning Generation}
\label{subsec:rise_cot}

\begin{figure*}[tb]
    \centering
    \includegraphics[width=\textwidth]{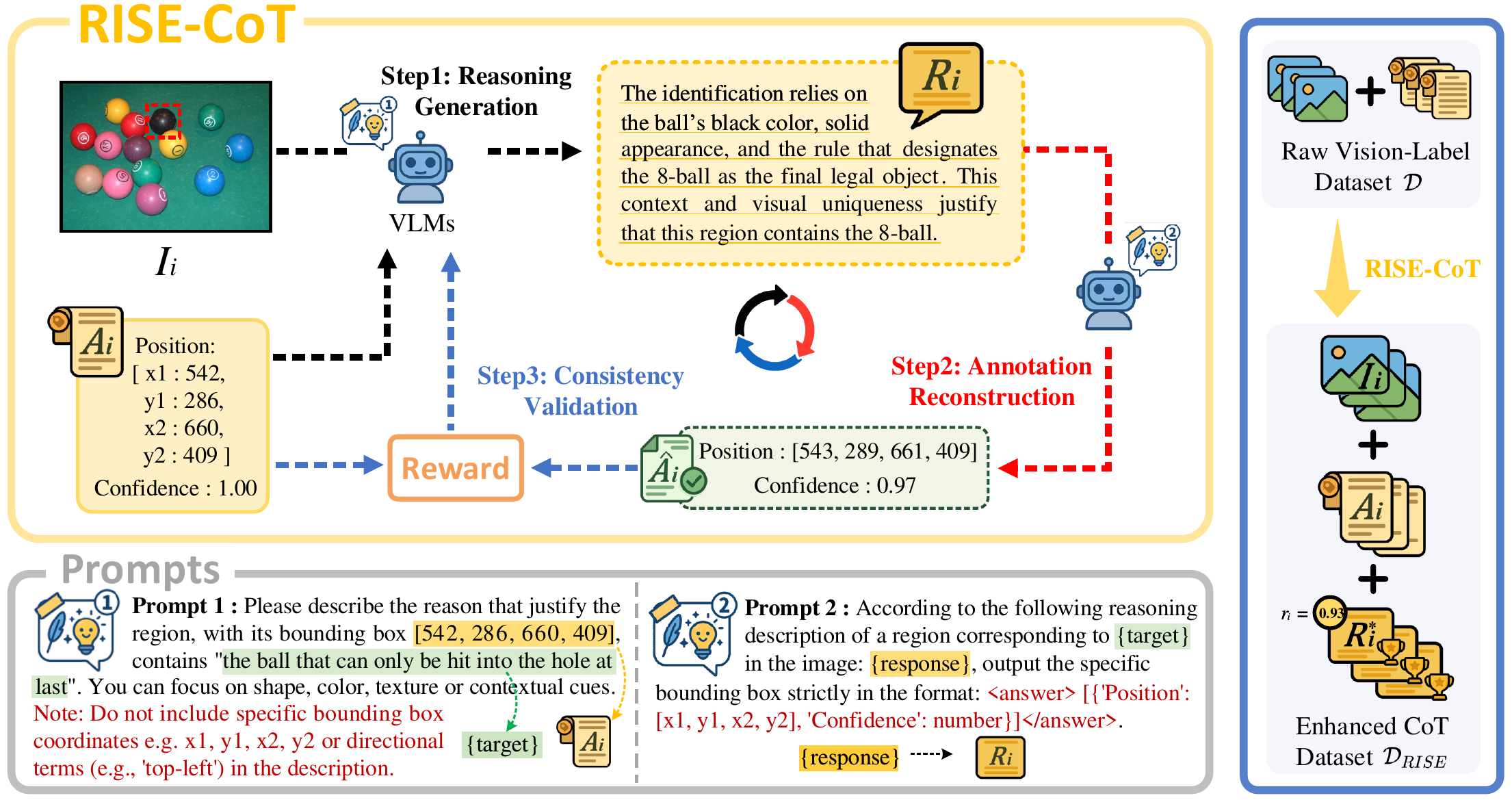}
    \captionsetup{style=ruled}
    \caption{RISE-CoT pipeline, the closed-loop process for producing high-quality CoTs from image-annotation pairs.}
    \label{fig:rise_cot}
\end{figure*}

RISE-CoT is the foundational component of our framework, tasked with generating high-quality, visually grounded CoTs for image-annotation pairs in a self-supervised manner. For a given dataset $\mathcal{D} = \{ (I_i, A_i) \}_{i=1}^M$, where $I_i$ is an image and $A_i$ is its corresponding annotation (e.g., probability distributions for classification or bounding boxes for detection), RISE-CoT operates through a closed-loop pipeline containing \textbf{Reasoning Generation}, \textbf{Annotation Reconstruction}, and \textbf{Consistency Validation} (see Figure~\ref{fig:rise_cot}).

In \textbf{Reasoning Generation}, a VLM produces a CoT $R_i$ that justifies the given annotation $A_i$, conditioned on both the image $I_i$ and $A_i$. The CoT is prompted to describe \textbf{visual and contextual cues without directly leaking annotation specifics}. For classification tasks, the CoT justifies the probability distribution, avoiding probability values. For detection tasks, the CoT justifies the bounding box, avoiding coordinates. This constraint prevents the VLM from simply regurgitating the annotation, compelling genuine, step-by-step reasoning based on visual evidence. We design prompts to explicitly instruct the VLM to describe visual cues (e.g., ‘smiling faces, festive settings’) rather than annotation specifics (e.g., ‘joy: 0.8’ or ‘[x, y, w, h]’), with template and examples provided \textbf{in Appendix A}.

In \textbf{Annotation Reconstruction}, a VLM reconstructs the annotation $\hat{A}_i$ from the generated CoT $R_i$ and image $I_i$ alone, verifying the CoT's quality. For classification, it predicts a probability distribution with $\sum p_i \approx 1$ enforced during training. For detection, it outputs bounding box coordinates in $[x, y, w, h]$ format. This ensures $R_i$ encapsulates sufficient visual and contextual cues to derive $A_i$ accurately. Prompt templates are also presented \textbf{in Appendix A}.

Carefully designed prompts are crucial for compelling the VLMs to produce visually grounded and logically consistent reasoning, ultimately enabling the objective evaluation of CoT quality. Figure~\ref{fig:rise_cot} presents examples of the \textbf{prompt structures and expected outputs}. This figure visually clarifies the relationship between the input image ($I_i$), the ground-truth annotation ($A_i$), the generated CoT ($R_i$), and the reconstructed annotation ($\hat{A}_i$). 

The final step, \textbf{Consistency Validation}, defines a reward function \(\mathcal{R}\) to ensure the generated reasoning's independence, format compliance, and reconstruction accuracy. The function, \textbf{with values ranging from 0 to 1}, is given by:
{\scriptsize
\begin{equation}
\mathcal{R}(A_i, \hat{A}_i, R_i) = S(A_i, \hat{A}_i) \cdot \{L(R_i, A_i) = 0 \wedge F_{CoT}(R_i, \hat{A}_i) = 1\},
\end{equation}
}
where \(L(R_i, A_i) = 0\) ensures no leakage of annotation specifics by checking for numerical content (probabilities or coordinate values) in \(R_i\), and \(F_{CoT}(R_i, \hat{A}_i) = 1\) verifies that \(R_i\) is a descriptive narrative and \(\hat{A}_i\) adheres to the task-specific format (e.g., probability distribution for classification, \([x, y, w, h]\) for detection. Moreover, the annotation similarity \(S(A_i, \hat{A}_i)\) quantifies how well the reconstructed annotation $\hat{A}_i$ matches the ground-truth annotation $A_i$, with task-specific definitions as follows:
\begin{itemize}
\item \textbf{For Classification}: \(S(A_i, \hat{A}_i)\) compares probability distributions for the \(i\)-th sample:
{\scriptsize
\begin{align}
S(A_i, \hat{A}_i) &= \phi(A_i, \hat{A}_i) \cdot \exp\left(-\left|\log_{10}\left(\sum_{j=1}^N \hat{A}_i(j)\right)\right|\right), \\
\phi(A_i, \hat{A}_i) &= \exp\left(-\left(KLD(A_i, \hat{A}_i) + MSE(A_i, \hat{A}_i)\right)\right), \\
\label{equ_kld}
KLD(A_i, \hat{A}_i) &= \sum_{j=1}^N A_i(j) \log \frac{A_i(j)}{\hat{A}_i(j)}, \\
MSE(A_i, \hat{A}_i) &= \frac{1}{N} \sum_{j=1}^N \left(A_i(j) - \hat{A}_i(j)\right)^2,
\end{align}
}
where \(A_i(j)\) and \(\hat{A}_i(j)\) are the ground-truth and reconstructed probabilities for the \(j\)-th class, respectively, across \(N\) classes (e.g., \(N=7\) for emotion classification). \(KLD(A_i, \hat{A}_i)\) is the Kullback-Leibler divergence, and \(MSE(A_i, \hat{A}_i)\) is the mean squared error, with zero values set to \(1e-10\). The term \(\exp\left(-\left|\log_{10}\left(\sum_{j=1}^N \hat{A}_i(j)\right)\right|\right)\) regularizes the probability sum to \(\approx 1\).

\item \textbf{For Detection}: \(S(A_i, \hat{A}_i)\) uses the average Intersection over Union (IoU) metric, computed after optimal matching of ground-truth and predicted bounding boxes using the Hungarian algorithm:
\begin{align}
\label{equ:S}
S(A_i, \hat{A}_i) &= \frac{1}{N_g} \sum_{j=1}^{N_g} \text{IoU}(A_{i,j}, \hat{A}_{i,k_j}),
\end{align}
where \(\text{IoU}(A_{i,j}, \hat{A}_{i,k_j})\) measures the overlap between the \(j\)-th ground-truth bounding box and its matched predicted bounding box \(k_j\) for image \(i\), determined by the Hungarian algorithm to maximize IoU. \(N_g\) is the number of ground-truth bounding boxes, and unmatched ground-truth boxes are assigned an IoU of 0. The average IoU, with denominator \(N_g\), accounts for all ground-truth boxes, penalizing unmatched detections and providing a robust measure of localization accuracy.
\end{itemize}

By compelling $R_i$ to avoid directly leaking annotation specifics, and by using reconstruction accuracy as the core reward, $\mathcal{R}$ establishes an objective and verifiable metric for CoT quality. 
% This effectively addresses the long-standing challenge of reliably assessing reasoning without relying on expensive human-annotated CoT data or heuristic rules.

\subsubsection{RISE-CoT Training}
RISE-CoT is trained using GRPO~\cite{deepseek2024deepseekmath} to optimize the policy $\pi_\theta(R_\theta, \hat{A}_\theta | I_i, A_i)$ by maximizing the expected reward:
\begin{align}
\mathcal{J}(\theta) &= \mathbb{E}_{(I_i, A_i) \sim \mathcal{D}} \left[ \mathcal{R}(A_i, \hat{A}_i, R_i) \right].
\end{align}
For each sample $(I_i, A_i)$, multiple CoTs and annotations are generated to identify the most robust reasoning path. The highest-reward CoT $R_i^*$ and its corresponding reward $r_i = \mathcal{R}(A_i, \hat{A}_i, R_i^*)$ are retained to form the high-quality CoT dataset $\mathcal{D}_\text{RISE} = \{ (I_i, R_i^*, A_i, r_i) \}_{i=1}^M$. \textbf{This dataset then serves as the training data for the subsequent stage}. This closed-loop design fundamentally aligns the generated reasoning $R_i$ with visual evidence, effectively compressing complex inputs into task-relevant, interpretable representations, adhering to the information bottleneck principle.

% Moreover, the output of RISE-CoT also enables annotation quality evaluation for simple tasks. High rewards indicate annotations consistent with visual evidence, while low rewards reveal errors (e.g., mislabeled categories), with reconstructed labels $\hat{A}_i$ suggesting corrections. This is validated in Section~\ref{sec:experiments} through noise detection and correction accuracy on ImageNet.

\subsection{RISE-R1: Training VLM for Enhanced CoTs}
\label{subsec:rise_r1}

\begin{figure}[tb]
\centering
\includegraphics[width=\linewidth]{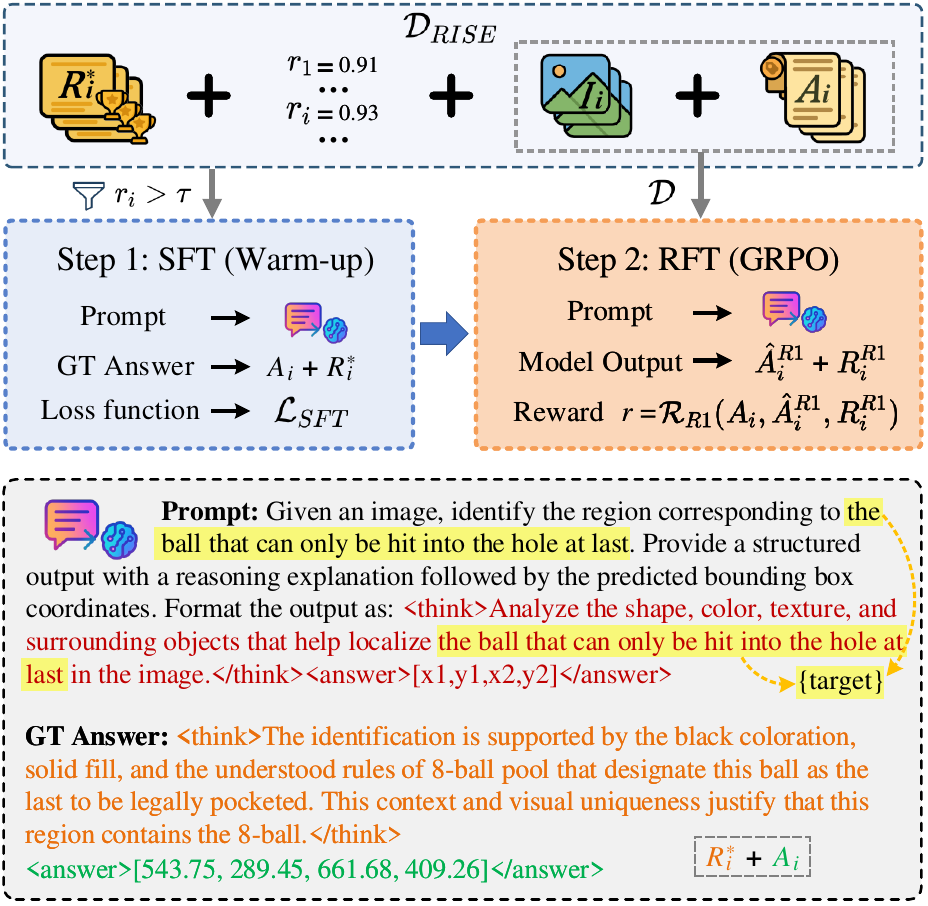}
\captionsetup{style=ruled}
\caption{RISE-R1 pipeline, a two-step training process involving SFT on a high-quality CoT subset and RFT to produce interpretable ``think-answer'' outputs.}
\label{fig:rise_r1}
\end{figure}

RISE-R1 trains a VLM to produce structured ``think-answer'' outputs, combining annotations \(\hat{A}_i\) and interpretable CoTs \(R_i\) from images. Following DeepSeek-R1’s two-stage approach~\cite{deepseek2023deepseekr1}, it initializes the VLM with SFT on a \textbf{high-quality CoT subset} from \(\mathcal{D}_\text{RISE}\) generated in RISE-CoT and further optimizes it using GRPO with RFT on full-scale \(\mathcal{D}\) as illustrated in Figure~\ref{fig:rise_r1}.

The dataset $\mathcal{D}_\text{RISE} = \{ (I_i, R_i^*, A_i, r_i) \}_{i=1}^M$ is constructed from the RISE-CoT stage, where $R_i^*$ denotes the highest-reward CoT and $r_i$ (the corresponding reward for image $I_i$). For the SFT step, we select a subset of high-quality CoTs, $\mathcal{D}_\text{RISE}^\text{high} = \{ (I_i, R_i^*, A_i) \mid r_i \geq \tau \}_{i=1}^K \subseteq \mathcal{D}_\text{RISE}$, where $\tau = 0.75$ (a threshold determined through ablation studies, \textbf{discussed in Appendix B}) ensures only CoTs with high reconstruction accuracy are utilized. Each pair $(I_i, R_i^*, A_i)$ forms a training sample where $(R_i^*, A_i)$ serves as the \textbf{ground-truth "think-answer" output} of $I_i$ for the VLM. Specifically, $R_i^*$ guides the ``think'' section and $A_i$ forms the ``answer'' section. The VLM is trained to minimize the standard supervised loss on $\mathcal{D}_\text{RISE}^\text{high}$. For each sample $(I_i, R_i^*, A_i) \in \mathcal{D}_\text{RISE}^\text{high}$, the VLM is prompted with only the input image $I_i$, including task descriptions and structured instructions to produce this logically organized "think-answer" output. Figure~\ref{fig:rise_r1} also demonstrates how the VLM is prompted with the input image and task instructions, and how these prompts are meticulously designed to guide the model in producing outputs that integrate the generated CoTs ($R_i^*$) and the original annotations $A_i$ into a structured "think-answer" format.

The loss function $\mathcal{L}_\text{SFT}$ is defined as:
\begin{align}
\mathcal{L}_\text{SFT} = -\frac{1}{K} \sum_{i=1}^K \log P_\theta(\text{`think-answer`}(R_i^*, A_i) | I_i),    
\end{align}
where $P_\theta(\text{`think-answer`}(R_i^*, A_i) | I_i)$ represents the joint probability of generating the complete "think-answer" sequence, which consists of the high-quality CoT $R_i^*$ followed by its annotation $A_i$, given the image $I_i$. This SFT pre-training phase instills the VLM with robust reasoning capabilities by directly learning from highly verifiable CoTs.

Following SFT, RISE-R1 employs RFT using GRPO~\cite{deepseek2023deepseekr1} on the full-scale dataset \(\mathcal{D} = \{ (I_i, A_i) \}_{i=1}^M\) to optimize the policy \(\pi_\theta(R_i, \hat{A}_i | I_i)\). In this stage, the VLM is prompted with the same 'question' input as in the SFT stage: the raw image $I_i$ along with general task instructions. This consistent input format ensures that the model learns to generate 'think-answer' outputs directly from the image during both training phases. Unlike RISE-CoT, which enforces no leakage in CoT generation, RISE-R1 focuses on enhancing the VLM’s end-to-end performance, producing precise \textbf{annotations} \(\hat{A}_i^{R1}\) and interpretable \textbf{CoTs} \(R_i^{R1}\) in the unified "think-answer" format. The reward function is defined as:
{\scriptsize
\begin{equation}
\mathcal{R}_\text{R1}(A_i, \hat{A}_i^{R1}, R_i^{R1}) = S(A_i, \hat{A}_i) \cdot \{F_{R1}(R_i^{R1}, \hat{A}_i^{R1}) = 1\},
\end{equation}
}
where \(S(A_i, \hat{A}_i^{R1})\) measures annotation similarity (the same in RISE-CoT) and \(F_{R1}(R_i^{R1}, \hat{A}_i^{R1}) = 1\) ensures format compliance, including the ''think-answer'' format (''\think{CoT}\answer{Annotation}'') and the annotation format (e.g., probability distributions for classification, \([x, y, w, h]\) for detection). The objective is to maximize the expected reward:
{
\begin{equation}
\mathcal{J}(\theta) = \mathbb{E}_{(I_i, A_i) \sim \mathcal{D}_\text{RISE}} \left[ \mathcal{R}_\text{R1}(A_i, \hat{A}_i^{R1}, R_i^{R1}) \right].
\end{equation}
}

During inference, only the RISE-R1-trained VLM is used to predict ``think-answer'' outputs from images. By pre-training on $\mathcal{D}_\text{RISE}^\text{high}$ and optimizing on $\mathcal{D}$, RISE-R1 outperforms Visual-RFT, as well as other methods. Ablation studies in Experiments Section validate the necessity of SFT pre-training with high-quality CoT subsets before RFT, and compare the impact of different CoT filtering thresholds.

\section{Experiments}
\label{sec:experiments}

\subsection{Experimental Setup}
\subsubsection{Datasets}
We evaluate RISE on four image annotation datasets varying levels of complexity and task demands.

\textbf{Emotion6} Classification:
The Cornell Emotion6 dataset~\cite{peng2015mixed} contains 1,980 images across 7 emotion categories: \{`\textit{anger}', `\textit{disgust}', `\textit{fear}', `\textit{joy}', `\textit{sadness}', `\textit{surprise}', `\textit{neutral}'\}. Ground-truth annotations are soft labels (probability distributions) derived from Amazon Mechanical Turk user ratings, normalized to sum to 1. It is split into 1,386 training (70\%) and 594 testing (30\%) images. Emotions are determined by complex contextual factors, such as scene settings, visual elements.

\textbf{LISA} Object Detection:
LISA Object Detection: The LISA dataset~\cite{lai2024lisa} is originally designed for reasoning-driven image segmentation with question-answer pairs (e.g., “\textit{In cold weather, dogs may need extra protection to keep them warm. What object in the picture can a dog wear to provide warmth during snowy walks?}” with a segmentation mask). We adapt it for object detection by: (1) extracting a single target per question (e.g., “\textit{a dog coat made of warm, water-resistant material and designed with cozy padding to keep a dog warm during snowy walks}”), (2) converting masks to bounding boxes, and (3) splitting images with multiple question-specific targets into single-target samples. This results in a dataset where the vast majority of samples contain a single target, with a total of 435 samples, split into 239 training and 196 testing samples. To simplify the task, we focus on single-target object detection for LISA, predicting one bounding box per image.

\textbf{ImageNet-Sub} Classification:  
We use a subset of the ImageNet~\cite{deng2009imagenet}, selecting 20 classes (\textbf{see Appendix C}). For each class, we randomly select 25 images for training and 10 for testing, totaling 500 training and 200 testing images. The single-label classification task requires outputting a probability distribution over all classes, with ground-truth as a one-hot distribution.

\textbf{COCO-Sub} Object Detection:
We use a subset of COCO~\cite{lin2014microsoft} (80 object classes). For each class, we randomly select 25 images for training and 10 for testing, resulting in 2,000 training and 800 testing images. Unlike LISA, COCO-Sub images often contain multiple targets per image, reflecting the dataset's natural complexity. For this task, we perform multi-target object detection, predicting all relevant bounding boxes per image to evaluate the model's ability to handle complex scenes with multiple objects.

\subsubsection{Implementation Details}
We fine-tune \textbf{Qwen2-VL-2B} on 8×NVIDIA RTX 4090 GPUs (32GB). Training uses AdamW (lr=$2\times10^{-5}$, weight decay=0.1), bf16 precision, and DeepSpeed ZeRO-3. We set: prompt length=1024, max image pixels=401{,}408, batch size=1, gradient accumulation=2, and total steps=200 (SFT/RFT). Flash Attention v2 and gradient checkpointing are enabled. For RISE-R1, we filter $\mathcal{D}_\text{RISE}^\text{high}$ using reward threshold $\tau=0.75$.

\subsubsection{Evaluation Metrics}
For \textbf{classification tasks}, we use \textbf{Jensen-Shannon divergence (JSD)} as the metric for Emotion6 classification, defined as 
\begin{align}
\label{equ_jsd}
    \text{JSD}(P, Q) = \frac{1}{2} \text{KLD}(P, M) + \frac{1}{2} \text{KLD}(Q, M),
\end{align}
with $M = \frac{P + Q}{2}$ and KLD is the Kullback-Leibler divergence. JSD’s symmetry and smoothing via the average distribution $M$ mitigate KLD’s sensitivity to extreme values, providing a more stable and representative measure of distribution alignment. For ImageNet-Sub, models predict a probability distribution over all classes, and we report \textbf{accuracy (Acc)} by selecting the class with the highest probability. For \textbf{detection tasks}, we use \textbf{mean Average Precision (mAP@0.5)} as the standard metric. 

\begin{table*}[htb]
    \centering
    \captionsetup{style=ruled}
    \caption{Comparisons on Complex Tasks. \textbf{Zero-shot} Base-Model and GPT-4o results are shown under Full-Shot.}
    \label{tab:emotion6_lisa_results}
    \begin{tabular}{l|cc|cc|cc|c|c|c}
        \hline
        \multirow{2}{*}{\textbf{Method}} & \multicolumn{6}{c|}{\textbf{Emotion6}} & \multicolumn{3}{c}{\textbf{LISA}} \\
        \cline{2-10}
        & \multicolumn{2}{c|}{\textbf{4-Shot}} & \multicolumn{2}{c|}{\textbf{16-Shot}} & \multicolumn{2}{c|}{\textbf{Full-Shot}} & \textbf{4-Shot} & \textbf{16-Shot} & \textbf{Full-Shot} \\
        \cline{2-10}
        & \textbf{JSD} & \textbf{WR} & \textbf{JSD} & \textbf{WR} & \textbf{JSD} & \textbf{WR} & \textbf{mAP} & \textbf{mAP} & \textbf{mAP} \\
        \hline
        Base-Model  & - & - & - & - & 0.470 &  0.67\% & - & - & 0.102 \\
        GPT-4o  & - & - & - & - & 0.106 & 34.01\% & - & - & 0.002 \\
        \hline
        SFT  & 0.220 & 6.23\% & 0.308 & 3.2\% & 0.112 & 31.14\% &  \textbf{0.200} & 0.204 & 0.367 \\
        Visual-RFT & 0.191 &  16.67\% & 0.222 &  12.12\% & 0.126 & 16.33\% & 0.130 & 0.246 & 0.395 \\
        \hline
        RISE & \textbf{0.168} &  & \textbf{0.133} &  & \textbf{0.071} &  & 0.195 & \textbf{0.271} & \textbf{0.404} \\
        \hline
    \end{tabular}
\end{table*}

\begin{table*}[htb]
    \centering
    \captionsetup{style=ruled}
    \caption{Comparisons on Simple Tasks. \textbf{Zero-shot} Base-Model and GPT-4o results are shown under Full-Shot.}
    \label{tab:imagenet_coco_results}
    \begin{tabular}{l|c|c|c|c|c|c}
        \hline
        \multirow{2}{*}{\textbf{Method}} & \multicolumn{3}{c|}{\textbf{ImageNet-Sub (Acc)}} & \multicolumn{3}{c|}{\textbf{COCO-Sub (mAP)}} \\
        \cline{2-7}
        & \textbf{4-Shot} & \textbf{16-Shot} & \textbf{Full-Shot} & \textbf{4-Shot} & \textbf{16-Shot} & \textbf{Full-Shot} \\
        \hline
        Base-Model  & - & - & 0.05 & - & - & 0.022 \\
        GPT-4o  & - & - & \textbf{0.925} & - & - & 0.000 \\
        \hline
        SFT  & \textbf{0.150} & 0.145 & 0.145 & 0.045 & 0.037 &  0.280 \\
        Visual-RFT & 0.040 &  0.090 & 0.100 & \textbf{0.101} & 0.029 & 0.284 \\
        \hline
        RISE & 0.120 & \textbf{0.155} & 0.545 & 0.087 & \textbf{0.094} & \textbf{0.302} \\
        \hline
    \end{tabular}
\end{table*}

\subsection{Results on Image Annotation Tasks}
We compare RISE with four baseline approaches including: (1) the standard Qwen2-VL-2B (\textbf{Base-Model}), evaluated in a \textbf{zero-shot} setting without fine-tuning; (2) \textbf{SFT}, a supervised fine-tuned variant of Base-Model trained to predict annotations $A_i$; (3) \textbf{Visual-RFT}~\cite{liu2025visual}, a RL-optimized variant of Base-Model trained under the same conditions as RISE-R1, representing RL-based methods like VLM-R1~\cite{shen2025vlm-r1}; and (4) \textbf{GPT-4o}, a state-of-the-art general VLM accessed via API and evaluated in a \textbf{zero-shot} setting.  Therefore, all approaches use \textbf{identical query prompts} (consistent with RISE-R1) to predict think-answer results, except SFT (only predict answer). 

\subsubsection{Complex Reasoning Tasks}
We evaluate RISE on Emotion6 and LISA, requiring complex reasoning (see Table~\ref{tab:emotion6_lisa_results}). For Emotion6, JSD measures the similarity between predicted annotations $\hat{A}^{R1}$ and ground truth $A$. We further report the \textbf{Win-Rate (WR)} for Emotion6, defined as the fraction of samples in which a baseline approach achieves a lower JSD than RISE. A higher WR thus indicates that the baseline outperforms RISE more frequently and is correspondingly stronger. For LISA, mAP@0.5 evaluates detection accuracy. In 4/16-shot settings, LISA uses 4/16 total samples, while Emotion6 uses 4/16 samples per class. As reported in~\cite{peng2015mixed}, a CNN model predicts probability distributions for Emotion6 with JSD worse than random ($JSD=0.22$) or uniform ($JSD=0.18$) distributions.

\subsubsection{Simple Straight Tasks}
We also evaluate RISE on ImageNet-Sub and COCO-Sub, representing simple tasks with direct visual-to-label mappings, using 4/16 samples per class in 4/16-shot settings. ImageNet-Sub is evaluated by accuracy (Acc). COCO-Sub is evaluated by mAP@0.5. The results are reported in Table~\ref{tab:imagenet_coco_results}.

Tables~\ref{tab:emotion6_lisa_results} and~\ref{tab:imagenet_coco_results} show that RISE perform well across complex and simple image annotation tasks. RISE excels in complex tasks by generating detailed CoTs and accurate annotations. However, in simple classification tasks like ImageNet-Sub, RISE's performance is constrained by the limited capacity of its base model, falling short of GPT-4o. The Base-Model, evaluated zero-shot, produces incorrect results due to its limited capability, but outperforms GPT-4o in detection tasks due to its \textbf{design optimized for object localization}~\cite{qwenteam2024qwen2vl}. SFT significantly improves Base-Model's task-specific performance. Visual-RFT, limited by suboptimal CoT WarmUp, shows slight gains over SFT, but lacks consistent improvement especially for complex reasoning tasks. GPT-4o, also zero-shot, achieves strong performance in qualitative classification tasks, especially simple tasks, but \textbf{fails in position-sensitive} detection tasks, highlighting its limitations in spatially precise annotations.

\begin{figure*}[tb]
    \centering
    \includegraphics[width=\textwidth]{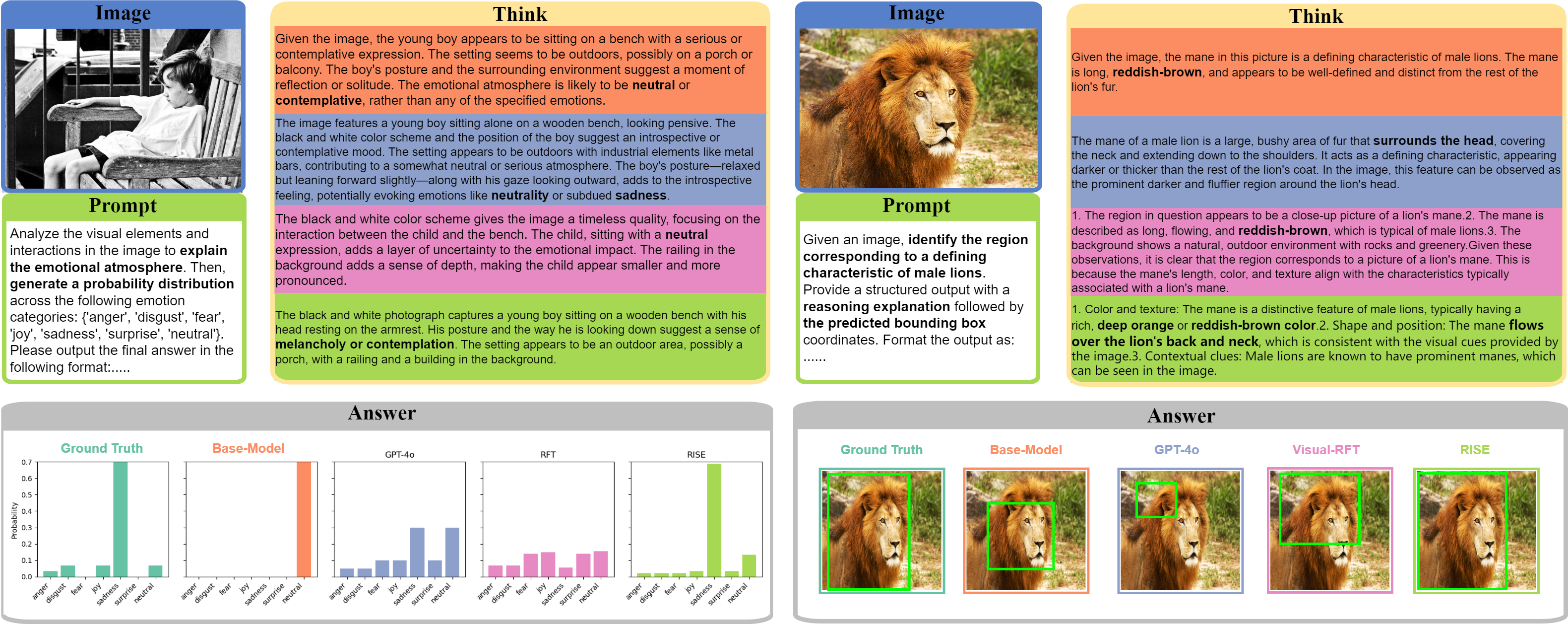}
    \captionsetup{style=ruled}
    \caption{Think-answer results of Emotion6 and LISA examples, with prompts partially shown due to space constraints.}
    \label{fig:comparison}
\end{figure*}

Figure~\ref{fig:comparison} provides the final ''think-answer'' comparisons of two examples, further demonstrating RISE's reasoning capabilities beyond quantitative metrics. \textbf{See more results in Appendix D}.

\subsection{Ablation Studies}
We conduct comprehensive ablation studies to dissect the contributions of key components within the RISE framework. Due to space constraints, we present results only for complex tasks (Emotion6 and LISA) in the ablation studies.

\subsubsection{CoTs Quality Evaluation of RISE-CoT}
We evaluate the quality of CoTs generated by RISE-CoT against two baselines: (1) \textbf{Base-Model-CoT} and (2) \textbf{GPT-4o-CoT}, where CoTs are respectively generated by QWen2-VL-2B and GPT-4o using the same Reasoning Generation prompts as RISE-CoT. For fair comparison, CoTs for all methods are generated on \(\mathcal{D}_\text{RISE}^\text{high}\), ensuring identical training samples. For each CoT source, we train RISE-R1 (SFT on \(\mathcal{D}_\text{RISE}^\text{high}\) with different CoTs followed by RFT on \(\mathcal{D}_\text{RISE}\)) and evaluate on Emotion6 and LISA. Table~\ref{tab:ablation_cot_quality} shows that RISE-CoT achieves the best performance across all datasets, demonstrating its ability to generate high-quality, visually grounded, and logically consistent CoTs.

\begin{table}[htb]
    \centering
    \captionsetup{style=ruled}
    \caption{Ablation on CoTs Quality in RISE-CoT.}
    \label{tab:ablation_cot_quality}
    \small
    \begin{tabular}{l|c|c}
        \toprule
        \textbf{Methods} & \textbf{Emotion6 (JSD)} & \textbf{LISA (mAP)} \\
        \midrule
        Base-Model-CoT & 0.120  & 0.311 \\
        GPT-4o-CoT & 0.098& 0.312 \\
        RISE-CoT & \textbf{0.071} & \textbf{0.404} \\
        \bottomrule
    \end{tabular}
\end{table}

% Figure~\ref{} gives some CoT examples generated by three approaches for Emotion6 and LISA respectively, clearly demonstrating RISE-CoT’s superior ability to elicit detailed and accurate reasoning CoTs, highlighting the strength of its annotation-reasoning-annotation closed loop.

\subsubsection{Role of SFT in RISE-R1}
We evaluate the design of RISE-R1 by comparing four training configurations: (1) \textbf{SFT Full}, only performing SFT on the full dataset \(\mathcal{D}_\text{RISE}\); (2) \textbf{SFT High}, only performing SFT on the high-quality subset \(\mathcal{D}_\text{RISE}^\text{high}\); (3) \textbf{RFT} (Visual-RFT\cite{liu2025visual}), only performing RFT on \(\mathcal{D}\); (4) \textbf{SFT Full+RFT}, performing SFT on \(\mathcal{D}_\text{RISE}\) followed by RFT on \(\mathcal{D}\); and (5) \textbf{SFT High+RFT}, performing SFT on \(\mathcal{D}_\text{RISE}^\text{high}\) followed by RFT on \(\mathcal{D}\). Table~\ref{tab:ablation_r1} shows that SFT High+RFT achieves the best performance, as SFT on high-quality CoTs provides critical initialization, enhancing RFT's optimization. SFT on \(\mathcal{D}_\text{RISE}\)+RFT performs slightly worse due to noisier CoTs in the full dataset, while SFT-only configurations underperform due to the lack of RFT's reinforcement learning.

\begin{table}[htb]
    \centering
    \captionsetup{style=ruled}
    \caption{Ablation on SFT and RFT in RISE-R1.}
    \label{tab:ablation_r1}
    \small
    \begin{tabular}{l|c|c}
        \toprule
        \textbf{Methods} & \textbf{Emotion6 (JSD)} & \textbf{LISA (mAP)} \\
        \midrule
        SFT Full & 0.094 & 0.354 \\
        SFT High & 0.085 & 0.391 \\
        RFT & 0.126 & 0.367 \\
        SFT Full+RFT & 0.102 & 0.365 \\
        SFT High+RFT & \textbf{0.071} & \textbf{0.404} \\
        \bottomrule
    \end{tabular}
\end{table}

\subsubsection{Reward Function Components}
We ablate the components of RISE-CoT's reward function \(\mathcal{R}\): (1) \textbf{Full \(\mathcal{R}\)}, including \(S(A_i, \hat{A}_i)\), \(P_l(R_i, A_i) = 0\), and \(M_f(R_i, \hat{A}_i) = 1\); (2) \textbf{Similarity-Only}, using only \(S(A_i, \hat{A}_i)\); (3) \textbf{No-Leakage-Removal}, omitting \(P_l(R_i, A_i) = 0\); and (4) \textbf{No-Format-Removal}, omitting \(M_f(R_i, \hat{A}_i) = 1\). Table~\ref{tab:ablation_reward} shows that the full \(\mathcal{R}\) achieves the best CoT quality and downstream performance, as leakage prevention ensures genuine reasoning and format constraints maintain output consistency.

\begin{table}[htb]
    \centering
    \captionsetup{style=ruled}
    \caption{Ablation on Reward Function Components.}
    \label{tab:ablation_reward}
    \small
    \begin{tabular}{l|c|c}
        \toprule
        \textbf{Methods} & \textbf{Emotion6 (JSD)} & \textbf{LISA (mAP)} \\
        \midrule
        Similarity-Only & 0.091 & 0.378 \\
        No-Leakage & 0.087 & 0.382 \\
        No-Format & 0.076 & 0.379 \\
        Full \(\mathcal{R}\) & \textbf{0.071} & \textbf{0.404} \\
        \bottomrule
    \end{tabular}
\end{table}

\subsubsection{Impact of Reward Threshold \(\tau\)}
The reward threshold \(\tau\) filters high-quality CoTs for \(\mathcal{D}_\text{RISE}^\text{high}\) in RISE-R1's SFT stage. We test \(\tau \in \{0.65, 0.75, 0.85\}\), with results in Table~\ref{tab:ablation_tau}. For Emotion6 and LISA, \(\tau = 0.75\) includes 41\% and 66\% of samples, respectively, balancing CoT quality and dataset size. Lower \(\tau = 0.65\) includes noisier CoTs, reducing performance, while higher \(\tau = 0.85\) limits samples, hindering generalization. Adjusting \(\tau\) for specific tasks is key to maintaining this balance. See more discussion in Appendix B.

\begin{table}[htb]
    \centering
    \captionsetup{style=ruled}
    \caption{Ablation on Reward Threshold \(\tau\).}
    \label{tab:ablation_tau}
    \small
    \begin{tabular}{l|c|c}
        \toprule
        \textbf{Methods} & \textbf{Emotion6 (JSD)} & \textbf{LISA (mAP)} \\
        \midrule
        \(\tau=0.65\) & 0.121 & 0.289 \\
        \(\tau=0.75\) & \textbf{0.071} & \textbf{0.404} \\
        \(\tau=0.85\) & 0.102 & 0.314 \\
        \bottomrule
    \end{tabular}
\end{table}

\section{Conclusion}
\label{sec:conclusion}
We introduced RISE, a novel two-stage framework that significantly enhances VLMs for complex image annotation tasks. In RISE-CoT stage, RISE autonomously generates high-quality CoTs by verifying their ability to reconstruct original annotations. These CoTs then power the RISE-R1 stage, training VLMs to produce accurate and interpretable "think-answer" outputs directly from images, eliminating the need for manually annotated CoT data.

Through its verifiable, self-supervised CoT generation, RISE improves annotation accuracy and interpretability while uniquely enabling implicit evaluation and refinement of dataset annotation quality. This framework effectively boosts the reasoning capabilities of lower-capacity VLMs across various image annotation tasks, allowing them to perform akin to larger models.

While RISE is designed for image annotation tasks where annotations are expressible as textual outputs and quantifiable, its applicability to tasks with non-textual annotations, tasks where correctness is harder to quantify, or common VQA tasks (not annotation-based) may require further adaptation. Future work will explore extending RISE to address these varied scenarios.

\bibliography{aaai2026}

\begin{thebibliography}{47}
\providecommand{\natexlab}[1]{#1}

\bibitem[{{Allen Institute for AI}(2024)}]{allen2024molmo}
{Allen Institute for AI}. 2024.
\newblock Molmo: Open-source multimodal models for efficient vision-language tasks.
\newblock \emph{arXiv preprint arXiv:2409.12345}.

\bibitem[{Azerbayev et~al.(2024)Azerbayev, Piotrowski, Assylbekov et~al.}]{deepseek2024deepseekmath}
Azerbayev, Z.; Piotrowski, B.; Assylbekov, Z.; et~al. 2024.
\newblock DeepSeekMath: Pushing the Limits of Mathematical Reasoning in Open Language Models.
\newblock \emph{arXiv preprint arXiv:2402.03300}.

\bibitem[{Baevski et~al.(2022)Baevski, Zhou, Mohamed, and Auli}]{baevski2022data2vec}
Baevski, A.; Zhou, Y.; Mohamed, A.; and Auli, M. 2022.
\newblock data2vec: A general framework for self-supervised learning in speech, vision and language.
\newblock \emph{arXiv preprint arXiv:2202.03555}.

\bibitem[{Bai et~al.(2023)Bai, Bai, Chu, Cui, Dang, Deng, Fan, Ge, Han et~al.}]{bai2023qwen}
Bai, J.; Bai, S.; Chu, Y.; Cui, Z.; Dang, K.; Deng, X.; Fan, Y.; Ge, W.; Han, Y.; et~al. 2023.
\newblock Qwen-VL: A frontier large vision-language model with versatile abilities.
\newblock \emph{arXiv preprint arXiv:2308.12966}.

\bibitem[{Caron, Touvron, and Joulin(2024)}]{caron2024dinov2vl}
Caron, M.; Touvron, H.; and Joulin, A. 2024.
\newblock DINOv2-VL: Extending self-supervised learning to vision-language tasks.
\newblock \emph{arXiv preprint arXiv:2403.12345}.

\bibitem[{Caron et~al.(2021)Caron, Touvron, Misra, J{\'e}gou, Mairal, Bojanowski, and Joulin}]{caron2021emerging}
Caron, M.; Touvron, H.; Misra, I.; J{\'e}gou, H.; Mairal, J.; Bojanowski, P.; and Joulin, A. 2021.
\newblock Emerging properties in self-supervised vision transformers.
\newblock \emph{arXiv preprint arXiv:2104.14294}.

\bibitem[{Chen et~al.(2020)Chen, Kornblith, Norouzi, and Hinton}]{chen2020simple}
Chen, T.; Kornblith, S.; Norouzi, M.; and Hinton, G. 2020.
\newblock A simple framework for contrastive learning of visual representations.
\newblock In \emph{International conference on machine learning}, 1597--1607. PMLR.

\bibitem[{{Deepseek Team}(2023)}]{deepseek2023deepseekr1}
{Deepseek Team}. 2023.
\newblock Deepseek-R1: Incentivizing reasoning capability in LLMs via reinforcement learning.
\newblock \emph{arXiv preprint arXiv:2311.12345}.

\bibitem[{Deng et~al.(2009)Deng, Dong, Socher, Li, Li, and Fei-Fei}]{deng2009imagenet}
Deng, J.; Dong, W.; Socher, R.; Li, L.-J.; Li, K.; and Fei-Fei, L. 2009.
\newblock {ImageNet: A Large-Scale Hierarchical Image Database}.
\newblock \emph{2009 IEEE Conference on Computer Vision and Pattern Recognition}, 248--255.

\bibitem[{Dosovitskiy et~al.(2021)Dosovitskiy, Beyer, Kolesnikov, Weissenborn, Zhai, Unterthiner, Dehghani, Minderer, Heigold, Gelly, Uszkoreit, and Houlsby}]{dosovitskiy2020image}
Dosovitskiy, A.; Beyer, L.; Kolesnikov, A.; Weissenborn, D.; Zhai, X.; Unterthiner, T.; Dehghani, M.; Minderer, M.; Heigold, G.; Gelly, S.; Uszkoreit, J.; and Houlsby, N. 2021.
\newblock An Image is Worth 16x16 Words: Transformers for Image Recognition at Scale.
\newblock In \emph{International Conference on Learning Representations}.

\bibitem[{{Google}(2023)}]{google2023gemini}
{Google}. 2023.
\newblock Gemini: A family of highly capable multimodal models.
\newblock Technical report, Google.

\bibitem[{{Google}(2024)}]{google2024gemini15}
{Google}. 2024.
\newblock Gemini 1.5: Unlocking multimodal understanding with long-context capabilities.
\newblock \emph{arXiv preprint arXiv:2402.12345}.

\bibitem[{Gupta, Dollar, and Girshick(2019)}]{gupta2019lvis}
Gupta, A.; Dollar, P.; and Girshick, R. 2019.
\newblock LVIS: A Dataset for Large Vocabulary Instance Segmentation.
\newblock In \emph{Proceedings of the IEEE/CVF Conference on Computer Vision and Pattern Recognition (CVPR)}, 5356--5364.

\bibitem[{He et~al.(2022)He, Chen, Xie, Li, Doll{\'a}r, and Girshick}]{he2022masked}
He, K.; Chen, X.; Xie, S.; Li, Y.; Doll{\'a}r, P.; and Girshick, R. 2022.
\newblock Masked autoencoders are scalable vision learners.
\newblock \emph{arXiv preprint arXiv:2111.06377}.

\bibitem[{He et~al.(2020)He, Fan, Wu, Xie, and Girshick}]{he2020momentum}
He, K.; Fan, H.; Wu, Y.; Xie, S.; and Girshick, R. 2020.
\newblock Momentum contrast for unsupervised visual representation learning.
\newblock \emph{arXiv preprint arXiv:1911.05722}.

\bibitem[{He et~al.(2015)He, Zhang, Ren, and Sun}]{he2016deep}
He, K.; Zhang, X.; Ren, S.; and Sun, J. 2015.
\newblock {Deep Residual Learning for Image Recognition}.
\newblock \emph{arXiv preprint arXiv:1512.03385}.
\newblock Available at \url{https://arxiv.org/abs/1512.03385}.

\bibitem[{Jiang et~al.(2025)Jiang, Zhang, Guo, Li, Qi, Chen, Wang, Jin, Guo, Yan et~al.}]{mme-cot2024}
Jiang, D.; Zhang, R.; Guo, Z.; Li, Y.; Qi, Y.; Chen, X.; Wang, L.; Jin, J.; Guo, C.; Yan, S.; et~al. 2025.
\newblock MME-CoT: Benchmarking Chain-of-Thought in Large Multimodal Models for Reasoning Quality, Robustness, and Efficiency.
\newblock \emph{arXiv preprint arXiv:2502.09621}.

\bibitem[{Kojima et~al.(2022)Kojima, Gu, Reid, Matsuo, and Iwasawa}]{kojima2022large}
Kojima, T.; Gu, S.~S.; Reid, M.; Matsuo, Y.; and Iwasawa, Y. 2022.
\newblock Large Language Models are Zero-Shot Reasoners.
\newblock \emph{CoRR}, abs/2205.11916.
\newblock ArXiv: 2205.11916.

\bibitem[{Lai et~al.(2024)Lai, Tian, Chen, Li, Yuan, Liu, and Jia}]{lai2024lisa}
Lai, X.; Tian, Z.; Chen, Y.; Li, Y.; Yuan, Y.; Liu, S.; and Jia, J. 2024.
\newblock Lisa: Reasoning segmentation via large language model.
\newblock In \emph{Proceedings of the IEEE/CVF Conference on Computer Vision and Pattern Recognition}, 9579--9589.

\bibitem[{Lam et~al.(2018)Lam, Kuzma, McGee, Dooley, Laielli, Klaric, Bulatov, and McCord}]{lam2018xview}
Lam, D.; Kuzma, R.; McGee, K.; Dooley, S.; Laielli, M.; Klaric, M.; Bulatov, Y.; and McCord, B. 2018.
\newblock xView: Objects in Context in Overhead Imagery.
\newblock In \emph{arXiv preprint arXiv:1802.07856}.

\bibitem[{LeCun et~al.(1989)LeCun, Boser, Denker, Henderson, Howard, Hubbard, and Jackel}]{lecun1989backpropagation}
LeCun, Y.; Boser, B.; Denker, J.~S.; Henderson, D.; Howard, R.~E.; Hubbard, W.; and Jackel, L.~D. 1989.
\newblock Backpropagation applied to handwritten zip code recognition.
\newblock \emph{Neural computation}, 1(4): 541--551.

\bibitem[{Li et~al.(2021)Li, Selvaraju, Gotmare, Joty, Xiong, and Hoi}]{li2021align}
Li, J.; Selvaraju, R.~R.; Gotmare, A.; Joty, S.; Xiong, C.; and Hoi, S. C.~H. 2021.
\newblock Align before fuse: Vision and language representation learning with momentum distillation.
\newblock \emph{Advances in neural information processing systems}, 34: 9694--9705.

\bibitem[{Li et~al.(2020{\natexlab{a}})Li, Wan, Cheng, Meng, and Han}]{li2020dior}
Li, K.; Wan, G.; Cheng, G.; Meng, L.; and Han, J. 2020{\natexlab{a}}.
\newblock DIOR: A large-scale benchmark dataset for remote sensing object detection.
\newblock \emph{ISPRS Journal of Photogrammetry and Remote Sensing}, 169: 374--386.

\bibitem[{Li et~al.(2020{\natexlab{b}})Li, Wan, Cheng, Meng, and Han}]{li2020remote}
Li, K.; Wan, G.; Cheng, G.; Meng, L.; and Han, J. 2020{\natexlab{b}}.
\newblock Object detection in optical remote sensing images: A survey and a new benchmark.
\newblock \emph{ISPRS Journal of Photogrammetry and Remote Sensing}, 159: 296--307.

\bibitem[{Lin et~al.(2014)Lin, Maire, Belongie, Hays, Perona, Ramanan, Doll{\'a}r, and Zitnick}]{lin2014microsoft}
Lin, T.-Y.; Maire, M.; Belongie, S.; Hays, J.; Perona, P.; Ramanan, D.; Doll{\'a}r, P.; and Zitnick, C.~L. 2014.
\newblock Microsoft COCO: Common objects in context.
\newblock In \emph{European conference on computer vision}, 740--755. Springer.

\bibitem[{Liu, Li, and Lee(2024)}]{liu2024llavaonevision}
Liu, H.; Li, C.; and Lee, Y.~J. 2024.
\newblock LLaVA-OneVision: Unified multi-image and video understanding.
\newblock \emph{arXiv preprint arXiv:2408.12345}.

\bibitem[{Liu et~al.(2023)Liu, Li, Wu, and Lee}]{liu2023visual}
Liu, H.; Li, C.; Wu, Q.; and Lee, Y.~J. 2023.
\newblock Visual instruction tuning.
\newblock \emph{arXiv preprint arXiv:2304.08485}.

\bibitem[{Liu et~al.(2025)Liu, Sun, Zang, Dong, Cao, Duan, Lin, and Wang}]{liu2025visual}
Liu, Z.; Sun, Z.; Zang, Y.; Dong, X.; Cao, Y.; Duan, H.; Lin, D.; and Wang, J. 2025.
\newblock Visual-RFT: Visual Reinforcement Fine-Tuning.
\newblock \emph{arXiv preprint arXiv:2503.01785}.

\bibitem[{{Mistral AI}(2024)}]{mistral2024pixtral}
{Mistral AI}. 2024.
\newblock Pixtral 12B: A lightweight multimodal model with native image processing.
\newblock \emph{arXiv preprint arXiv:2409.12346}.

\bibitem[{Mogelmose, Trivedi, and Moeslund(2012)}]{mogelmose2012vision}
Mogelmose, A.; Trivedi, M.~M.; and Moeslund, T.~B. 2012.
\newblock {Vision-Based Traffic Sign Detection and Analysis for Intelligent Driver Assistance Systems: Perspectives and Survey}.
\newblock \emph{IEEE Transactions on Intelligent Transportation Systems}, 13(4): 1484--1497.

\bibitem[{{OpenAI}(2023)}]{openai2023gpt4v}
{OpenAI}. 2023.
\newblock GPT-4v: Technical Report.
\newblock Technical report, OpenAI.

\bibitem[{Panda et~al.(2021)Panda, Pal, Banerjee, and Mitra}]{panda2021webemo}
Panda, R.; Pal, U.; Banerjee, A.; and Mitra, A. 2021.
\newblock {WebEmo: Understanding Emotions in Web Images through Multimodal Language Models}.
\newblock \emph{arXiv preprint arXiv:2112.09761}.
\newblock Available at \url{https://arxiv.org/abs/2112.09761}.

\bibitem[{Peng et~al.(2015)Peng, Chen, Sadovnik, and Gallagher}]{peng2015mixed}
Peng, K.-C.; Chen, T.; Sadovnik, A.; and Gallagher, A.~C. 2015.
\newblock A mixed bag of emotions: Model, predict, and transfer emotion distributions.
\newblock In \emph{Proceedings of the IEEE conference on computer vision and pattern recognition}, 860--868.

\bibitem[{{Qwen Team}(2024)}]{qwenteam2024qwen2vl}
{Qwen Team}. 2024.
\newblock Qwen2-VL: Advancing vision-language understanding with dynamic resolution and multimodal capabilities.
\newblock \emph{arXiv preprint arXiv:2409.12191}.

\bibitem[{Radford et~al.(2021)Radford, Kim, Hallacy, Ramesh, Goh, Agarwal, Sastry, Askell, Mishkin, Clark et~al.}]{radford2021learning}
Radford, A.; Kim, J.~W.; Hallacy, C.; Ramesh, A.; Goh, G.; Agarwal, S.; Sastry, G.; Askell, A.; Mishkin, P.; Clark, J.; et~al. 2021.
\newblock Learning transferable visual models from natural language supervision.
\newblock In \emph{International conference on machine learning}, 8748--8763. PMLR.

\bibitem[{Redmon et~al.(2016)Redmon, Divvala, Girshick, and Farhadi}]{redmon2016you}
Redmon, J.; Divvala, S.; Girshick, R.; and Farhadi, A. 2016.
\newblock You only look once: Unified, real-time object detection.
\newblock In \emph{Proceedings of the IEEE conference on computer vision and pattern recognition}, 779--788.

\bibitem[{Ronneberger, Fischer, and Brox(2015)}]{ronneberger2015u}
Ronneberger, O.; Fischer, P.; and Brox, T. 2015.
\newblock U-net: Convolutional networks for biomedical image segmentation.
\newblock In \emph{International Conference on Medical image computing and computer-assisted intervention}, 234--241. Springer.

\bibitem[{Schulman et~al.(2017)Schulman, Wolski, Dhariwal, Radford, and Klimov}]{schulman2017proximal}
Schulman, J.; Wolski, F.; Dhariwal, P.; Radford, A.; and Klimov, O. 2017.
\newblock Proximal policy optimization algorithms.
\newblock \emph{arXiv preprint arXiv:1707.06347}.

\bibitem[{Shen et~al.(2025)Shen, Liu, Li, Fang, Ma, Liao, Shen, Zhang, Zhao, Zhang et~al.}]{shen2025vlm-r1}
Shen, H.; Liu, P.; Li, J.; Fang, C.; Ma, Y.; Liao, J.; Shen, Q.; Zhang, Z.; Zhao, K.; Zhang, Q.; et~al. 2025.
\newblock Vlm-r1: A stable and generalizable r1-style large vision-language model.
\newblock \emph{arXiv preprint arXiv:2504.07615}.

\bibitem[{Wah et~al.(2011)Wah, Branson, Welinder, Perona, and Belongie}]{wah2011caltech}
Wah, C.; Branson, S.; Welinder, P.; Perona, P.; and Belongie, S. 2011.
\newblock The Caltech-UCSD Birds-200-2011 Dataset.
\newblock Technical Report CNS-TR-2011-001, California Institute of Technology.

\bibitem[{Wei et~al.(2022)Wei, Wang, Schuurmans, Bosma, Xia, Chi, Le, Zhou et~al.}]{wei2022chain}
Wei, J.; Wang, X.; Schuurmans, D.; Bosma, M.; Xia, F.; Chi, E.~H.; Le, Q.~V.; Zhou, D.; et~al. 2022.
\newblock Chain-of-thought prompting elicits reasoning in large language models.
\newblock \emph{Advances in neural information processing systems}, 35: 24824--24837.

\bibitem[{Yao et~al.(2023)Yao, Yu, Zhao, Shafran, Griffiths, Cao, and Narasimhan}]{yao2023tree}
Yao, S.; Yu, D.; Zhao, J.; Shafran, I.; Griffiths, T.~L.; Cao, Y.; and Narasimhan, K. 2023.
\newblock Tree of Thoughts: Deliberate problem solving with large language models.
\newblock \emph{arXiv preprint arXiv:2305.10601}.

\bibitem[{Yao et~al.(2022)Yao, Zhao, Yu, Du, Shafran, Narasimhan, and Cao}]{yao2022react}
Yao, S.; Zhao, J.; Yu, D.; Du, N.; Shafran, I.; Narasimhan, K.; and Cao, Y. 2022.
\newblock ReAct: Synergizing reasoning and acting in language models.
\newblock \emph{arXiv preprint arXiv:2210.03629}.

\bibitem[{Zhang et~al.(2022)Zhang, Xie, Xia, and Shen}]{zhang2022medical}
Zhang, J.; Xie, Y.; Xia, Y.; and Shen, C. 2022.
\newblock Medical image segmentation using deep learning: A survey.
\newblock \emph{arXiv preprint arXiv:2009.13120}.

\bibitem[{Zhang et~al.(2023)Zhang, Zhang, Li, and Smola}]{zhang2023automatic}
Zhang, Z.; Zhang, A.; Li, M.; and Smola, A. 2023.
\newblock Automatic Chain of Thought Prompting in Large Language Models.
\newblock In \emph{International Conference on Learning Representations (ICLR)}.
\newblock ArXiv: 2210.03493.

\bibitem[{Zhou et~al.(2022)Zhou, Wei, Wang, Shen, Xie, Yuille, and Kong}]{zhou2022ibot}
Zhou, J.; Wei, C.; Wang, H.; Shen, W.; Xie, C.; Yuille, A.; and Kong, T. 2022.
\newblock iBOT: Image BERT pre-training with online tokenizer.
\newblock \emph{arXiv preprint arXiv:2111.07832}.

\bibitem[{Zoph et~al.(2020)Zoph, Ghiasi, Lin, Cui, Liu, Cubuk, and Le}]{zoph2020rethinking}
Zoph, B.; Ghiasi, G.; Lin, T.-Y.; Cui, Y.; Liu, H.; Cubuk, E.~D.; and Le, Q.~V. 2020.
\newblock Rethinking pre-training and self-training.
\newblock In \emph{Advances in neural information processing systems}, volume~33, 19186--19197.

\end{thebibliography}
% 导入整个 PDF


\section*{Supplementary material}

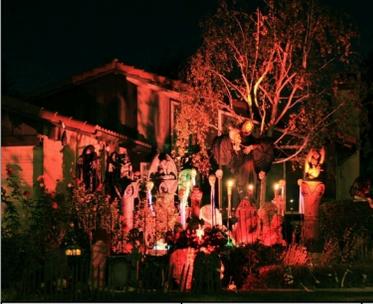

| | **Prob Distribution** |
|---|---|
| | {'anger': 0.0, 'disgust': 0.1, 'fear': 0.2, 'joy': 0.388889, 'sadness': 0.033333, 'surprise': 0.122222, 'neutral': 0.155556} |

| STEP | Reward | COT |
|---|---|---|
| **0** | **0.643** | In the image, we observe a festive and eerie display of decorations surrounding a house at night. The decorations include various statues and figures that suggest themes of horror and mystery. Red lighting illuminates the scene, creating a stark contrast against the dark surroundings. The illuminated statues appear to be looking off into the distance, adding to the eerie ambiance. The presence of tall candles further emphasizes the somber mood. The overall interpretation of emotions based on this detailed description is likely:Fear: Given the presence of menacing-looking statues and the overall unsettling atmosphere created by the vivid lighting.These interpretations are derived from typical associations commonly associated with Halloween decorations, where horror stories and eerie themes are often emphasized. |
| **100** | **0.774** | The image depicts a house adorned with Halloween decorations, illuminated with red lights. The house features a large, tree-shaped display made up of numerous creepy and whimsical figures, including skeletons and creatures with intricate facial features. There is a mix of emotions depicted in the scene - the vivid red light casts an intense red hue over the entire structure, which could be perceived as both delightful and unsettling. The figure of an angry or menacing character among the displays could imply a negative emotional response, contrasting sharply with the lightness of some of the decorations. The overall atmosphere is one of mixed emotions, with a strong presence of horror and disapproval due to the aggressive or menacing figures, while the whimsical and colorful decorations provide a counterbalance of mirth and amusement. |
| **200** | **0.950** | The image depicts a Halloween house decorated with elaborate Halloween decorations illuminated by lights. The environment is dark with subtle hints of darkness surrounding the illuminated area around the center. Key elements include large, illuminated figures resembling ghosts or monsters, some holding candles. The setting appears festive but somber due to the overall darkness.Surprise: Low – There isn't anything specifically surprising about the image; it's more about the arrangement of objects than any immediate actions. It's more likely a festive activity rather than a situation with personal loss. |

| Dimension | CoT 1 (Reward: 0.643) | CoT 2 (Reward: 0.774) | CoT 3 (Reward: 0.950) |
|---|---|---|---|
| **Structure & Clarity** | ❌ Verbose and somewhat disjointed; conclusion feels bolted on at the end | ✅ Clear flow of ideas with transitions from observation to emotion | ✅ Very concise, step-by-step reasoning; flows logically and efficiently from setting to emotional interpretation |
| **Visual Detail Usage** | ✅ Includes specific elements (statues, red lights, candles), but some are repetitive or generic | ✅ Rich visual descriptions (e.g., "tree-shaped display," "skeletons," "creatures with facial features") | ✅ Selectively mentions key features (ghosts/monsters, candles) with a focus on what's emotionally relevant |
| **Emotional Reasoning** | ❌ Limited to one emotion ("fear"); reasoning is generic and relies on stereotypical associations | ✅ Highlights emotional complexity — blends "delightful" with "unsettling," contrasts menace vs. whimsy | ✅ Refines emotional judgment — interprets mood as "festive but somber," and evaluates low *surprise* with logical justification |
| **Contextual Awareness** | ✅ Refers to "Halloween" and common associations with horror | ✅ Builds deeper context with contrasts (aggressive vs. whimsical), indicating nuanced understanding of the scene | ✅ Applies functional context ("festive activity," "not personal loss") and links setting to emotion subtly |
| **Conclusion Strength** | ❌ Weak – overly reliant on assumptions about Halloween rather than grounded evidence | ✅ Balanced emotional conclusion grounded in visual cues and thematic contrast | ✅ Strong, precise conclusion — avoids exaggeration, uses "surprise: low" as an analytical statement rather than a guess |

Figure 8: CoT evolution for sample images in Emotion6 during RISE-CoT training, showing CoT descriptions and reward scores at steps 0, 100, and 200.

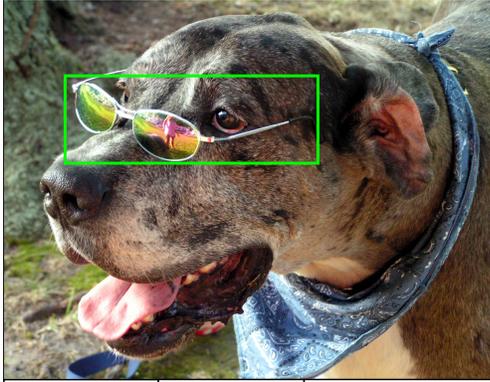

| | Target Description |
| --- | --- |
| | **The object in the picture that could be used to block the sun's glare and protect the eyes.** |

| STEP | Reward | COT |
| --- | --- | --- |
| **0** | **0.406** | 1. The image shows a region with window features, including rectangular windows and a sloped design typically associated with roof windows.2. The window is in a room, not a building exterior.3. The environment suggests it was probably installed or being prepared for installation, as indicated by the ladder and other materials.4. The windows provide natural light, which is common in windows.Given these observations, the region corresponds to a window because it is rectangular, designed to admit light, and fits within a room setting and context. |
| **100** | **0.536** | 1. Type of object: The image shows a window.2. Natural lighting feature: The window is designed to let in natural light, aligning with the region's role in providing illumination inside the room.3. Shape and configuration: The window has a typical rectangular shape and is set into framed beams, showing it is not a skylight.4. Contextual relevance: The region is a window because it has been mentioned as providing natural light and it is likely set in a room given the presence of other room elements.5. Conclusion: Based on these observations, the region corresponds to the target object, a window, due to its shape, lighting purpose, and contextual similarity to a window's appearance.Chain of Thought:1. Identifying the shape typical of windows for natural lighting.2. Identifying the natural-light-giving role of the window.3. Integrating the contextual elements such as other room components and roofline.4.Concluding the observation as a window based on its visual characteristics. |
| **200** | **0.834** | 1. Observe the shape and size: The window has a square or rectangular shape and is mounted on the wall, which matches the characteristics of typical window shapes.2. Identify any indications of glass: There are no visible pieces of glass seen in this region, which further supports the identification that it is a window.3. Compare with known objects: Common windows are typically rectangular or square and can be opened or closed to allow light and air into a room.4. Context: The room has walls and flooring that are clean and unpainted, suggesting that this might be a new construction area.Based on these observations, the region corresponding to the target object (a window) is confirmed because of the square or rectangular shape typical of windows, the absence of glass, and the context of the room being under construction or refurbishment. |

| Dimension | CoT 1 (Reward: 0.240) | CoT 2 (Reward: 0.626) | CoT 3 (Reward: 0.834) |
| --- | --- | --- | --- |
| **Structure & Clarity** | ❌ Uses generic labels like "Cognitive Processing"; lacks a tight reasoning chain | ✅ Step-by-step structure with clear visual and contextual progression | ✅ ✅ Highly structured reasoning steps with logical flow and summary |
| **Visual Feature Analysis** | ❌ Basic mention of shape and texture; lacks concrete visual grounding | ✅ Analyzes shape, color, texture in more detail (e.g., rolled hem, shade of blue) | ✅ ✅ Mentions specific visual cues (e.g., reflection in glasses, worn around the neck) that are directly relevant |
| **Contextual Integration** | ❌ Minimal context use (e.g., says it's for comfort but doesn't support it well) | ✅ Considers usage (blocking sun, worn as headband), better aligned with function | ✅ ✅ Strong context use (mentions reflection in glasses, dog's appearance), reinforces object identity |
| **Semantic Accuracy** | ❌ Overstates practicality ("designed for protective use") without evidence | ✅ Reasonable assumptions about function (sun glare protection, headband) | ✅ ✅ Function is well justified through visual + contextual synthesis |
| **Conclusion Strength** | ❌ Vague phrasing: "most likely a bandana" | ✅ Clearer and supported conclusion | ✅ ✅ Strong and confident conclusion, tightly linked to all evidence presented |

Figure 9: CoT evolution for sample images in LISA during RISE-CoT training, showing CoT descriptions and reward scores at steps 0, 100, and 200.

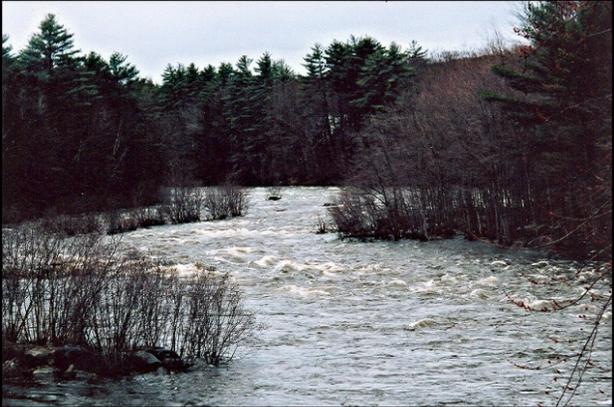

| | **Prob Distribution** |
|---|---|
| | {'anger': 0.0333333, 'disgust': 0.0, 'fear': 0.266667, 'joy': 0.4, 'sadness': 0.0, 'surprise': 0.0333333, 'neutral': 0.266667} |
| **Reward** | **COT** |
| **0.077** | Here are the key objects, actions, and expressions observed from the image:1.2. Tree Lines: The line of trees flanking the river adds a sense of stability and calm, contrasting with the dynamic nature of the river.3.4. Cloudy Sky: The overcast sky can contribute to a sense of uncertainty or gloominess, which might contrast with the chaos of the river and reinforce the overall emotions.5. Water Depth: The varying depths of the river suggest movement and change, which can be interpreted as the unpredictable nature of emotions.Based on these observations:- Neutral (Neutral): The combination of serene forest setting, calm river flow, and cloudy sky creates a neutral atmosphere |

### Failure Analysis of the Chain-of-Thought

**1.Confusing and incomplete structure:** The numbering is unclear and some points are missing, disrupting the logical flow.
**2.Vague descriptions:** Key elements like "Tree Lines" and "Cloudy Sky" lack detailed visual observations.
**3.Shallow reasoning:** The conclusion of a "neutral" atmosphere is given without strong justification or clear links to the observations.
**4.Repetitive language:** Similar ideas are repeated without clear progression or deeper insight.
**5.Lack of causal connections:** Observations are fragmented and not well connected, leading to unsupported conclusions.

Figure 10: Failure cases in RISE-CoT for Emotion6, showing original image descriptions, Ground Truth annotations, reconstructed annotations, and low reward scores due to unclear visual cues.

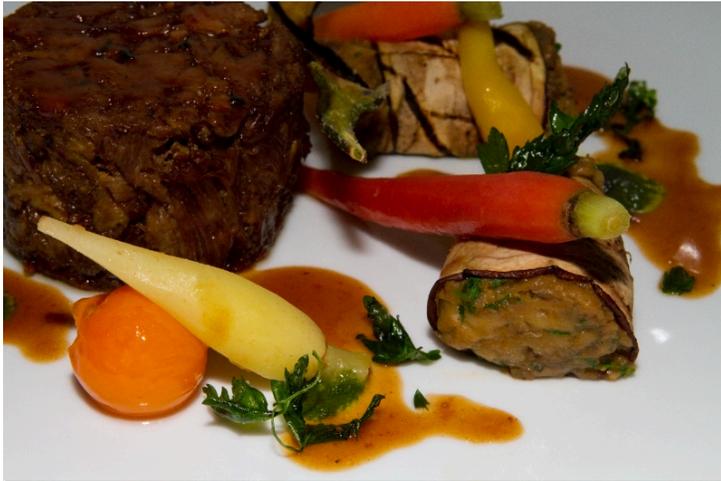

| Reward | COT |
|--------|-----|
| | **Target Description** |
| | The object which medical professionals wear during surgery to maintain hygiene and prevent contamination |
| **0.260** | 1. The object in question is identified as a slice of bacon due to its characteristic shape and texture.2. The image shows the slice of bacon with a thin, elongated shape and a slightly curly edge, which resembles the common profile of bacon.3. The texture of the slice appears to be crispy and well-flavored, which is typical of bacon.4. The surrounding context, such as the beef steak and vegetables, is also consistent with traditional bacon pairing, where bacon is used to enhance the flavor of the steak.5. These visual cues and contextual information confirm that the object located at is a slice of bacon.Therefore, the final answer is that the region corresponds to a slice of bacon because it meets the criterion of a thin, elongated, crispy, well-flavored, often used meat to enhance the taste of other dishes. |

**Failure Analysis of the Chain-of-Thought**

**1.Lack of visual grounding:** The CoT repeatedly uses subjective terms like "crispy" and "well-flavored", which cannot be directly observed in the image. These are speculative descriptions rather than visual evidence.
**2.Weak contextual integration:** Although it mentions "beef steak and vegetables," it relies on general food pairings rather than analyzing the actual positioning or interaction of objects in the image.
**3.Redundant and repetitive reasoning:** Multiple steps restate the same point—that the object is bacon due to its shape and texture—without adding new layers of reasoning, making the logic feel flat and circular.
**4.Grammatical and structural issues:** The sentence "the object located at is a slice of bacon" contains an obvious placeholder or formatting error, hurting clarity and professionalism.
**5.Descriptive definition over analytical reasoning:** The conclusion focuses on what bacon is ("a thin, elongated, well-flavored meat used to enhance dishes") instead of why the object in the image specifically matches those features.

Figure 11: Failure cases in RISE-CoT for LISA, showing original image descriptions, Ground Truth annotations, reconstructed annotations, and low reward scores due to unclear visual cues.



\end{document}

\begin{figure*}[hbtp]
    \centering
    \includegraphics[width=\linewidth]{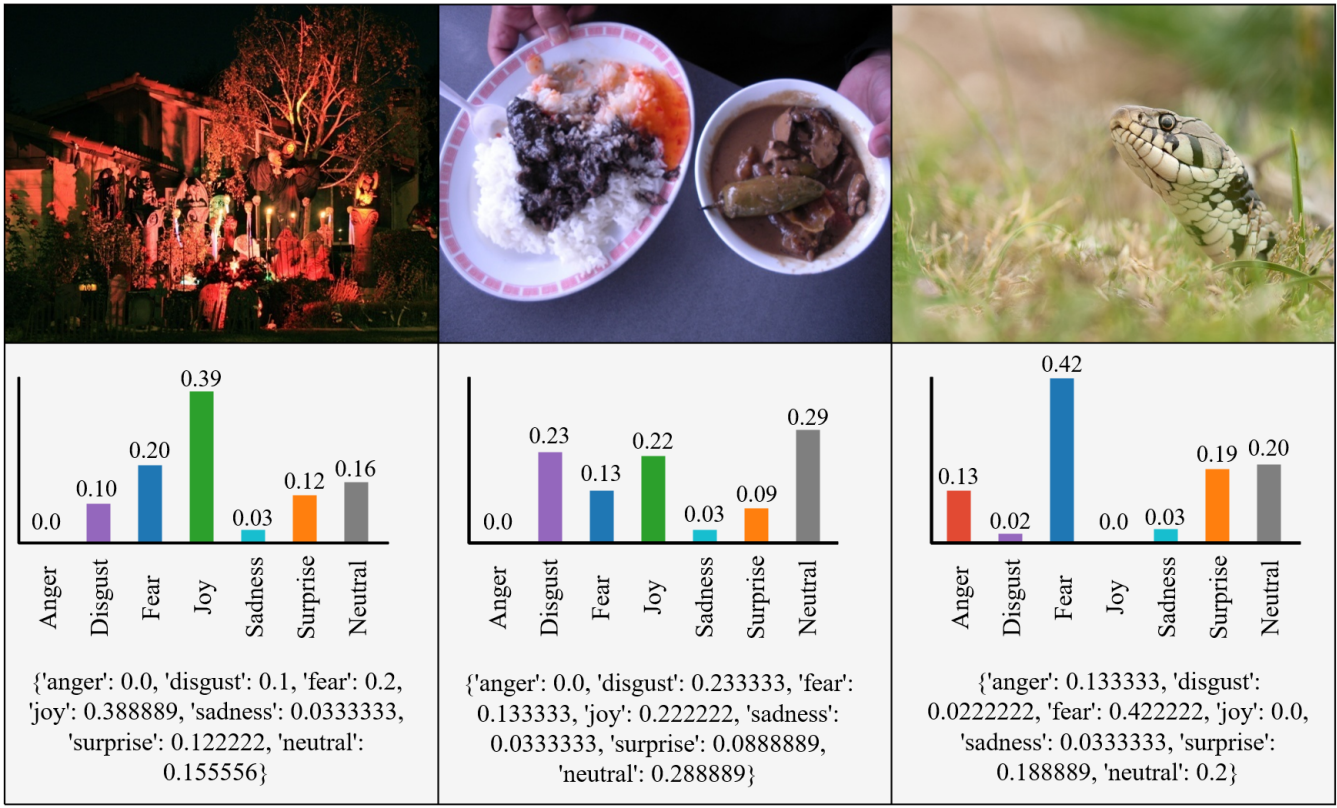}
    \captionsetup{style=ruled}
    \caption{Example image from Emotion6 and Ground Truth classification annotation, showing a probability distribution over six emotion categories.}
    \label{fig:emotion6_example}
\end{figure*}

\appendix
\section{Appendix A: Prompt Templates}
\label{sec:appendix_a}
Below, we provide the prompt templates of RISE-CoT and RISE-R1 stages for both classification (Emotion6) and detection (LISA) tasks, along with example inputs and expected outputs.

\subsection{Dataset and Task Overview}

% Introducing Emotion6 dataset and task
The Emotion6\cite{peng2015mixed} dataset is designed for emotion classification, where images are labeled with a probability distribution across six emotion categories: anger, disgust, fear, joy, sadness, and neutral. The classification task requires analyzing visual elements, such as objects, colors, and textures, to assign probabilities to each emotion category, reflecting the image’s emotional atmosphere. Figure~\ref{fig:emotion6_example} provides an example image description and its Ground Truth classification annotation to illustrate the task.

% Introducing LISA dataset and task
The LISA dataset, originally a Visual Question Answering (VQA) dataset\cite{lai2024lisa} with Mask annotations, was adapted for target detection. We converted the original questions into Target Descriptions and transformed the Mask annotations into Bounding Boxes (BBox), creating a Target Detection dataset. The detection task involves identifying and localizing the target object based on its description, focusing on visual cues like shape, color, texture, and context. Figure~\ref{fig:lisa_example} provides an example image description, the original VQA question with its Mask answer, and the converted Target Description with its corresponding BBox used for RISE-CoT prompt design.

\begin{figure*}[htbp]
    \centering
    \includegraphics[width=\linewidth]{LaTeX/ExampleLISA.png}
    \captionsetup{style=ruled}
    \caption{Example image from LISA with its description, original VQA question and Mask answer, and converted Target Description with Bounding Box for target detection.}
    \label{fig:lisa_example}
\end{figure*}

\subsubsection{Reasoning Generation Prompts}

The Reasoning Generation step prompts the VLM to produce a CoT ($R_i$) that justifies the annotation ($A_i$) based on visual and contextual cues, without leaking annotation specifics. The prompts are structured as follows:

\begin{itemize}
\item \textbf{Classification (Emotion6)}:
\textit{Analyze the image to explain how the visual elements and interactions support the emotional probability distribution \textbf{\$prob\_distribution\$}, concisely in under 100 words. Describe key objects, actions, and expressions that contribute to the inferred emotional atmosphere, highlighting their connection to the emotional probabilities. Avoid directly stating the emotions or probabilities; instead, focus on the visual cues and their implications.}

\item \textbf{Detection (LISA)}:
% \textit{Please describe the reason that justifies the region, with its bounding box \textbf{\$[x1,y1,x2,y2]\$}, contains \textbf{\$target\$}. You can focus on shape, color, texture, or contextual cues. Note: Do not include specific bounding box coordinates or directional references in the description.}
\textit{Please describe the reason that justify the region, with its bounding box \textbf{\$[x1, y1, x2, y2]\$}, contains \textbf{\$target\$}. You can focus on shape,color,texture or contextual cues. Note: Do not include specifc bounding box coordinates e.g. x1, y1, x2, y2 or directional terms (e.g., 'top-left') in the description. }
\end{itemize}

\begin{table}[htbp]
\centering
\small
\setlength{\tabcolsep}{3pt}     
\renewcommand{\arraystretch}{1.1}
\begin{tabularx}{\columnwidth}{@{}>{\raggedright}X>{\raggedright}X>{\raggedright\arraybackslash}X@{}}
\toprule
\textbf{Task} & \textbf{Template} & \textbf{Example} \\ \midrule
Classification \newline (Emotion6) & \codecell{prob\_distribution} & \codecell{\{'anger': 0.0,\newline
 'disgust': 0.1,\newline
 'fear': 0.2,\newline
 'joy': 0.388889,\newline
 'sadness': 0.033333,\newline
 'surprise': 0.122222,\newline
 'neutral': 0.155556\}} \\[2pt] \hline
Detection\newline (LISA) & \codecell{[x1,y1,x2,y2]}  & \codecell{[138, 182, 656, 428]} \\[2pt] \hline
Detection\newline (LISA) & \codecell{target} & \codecell{The object in the picture that could be used to block the sun's glare and protect the eyes} \\
\bottomrule
\end{tabularx}
\caption{RISE-CoT Template Variables and Examples.The examples are drawn from the first of the three targets in Figure 1 and Figure 2 respectively.}
\label{tab:template_vars}
\end{table}

\begin{table*}[t]
\centering
\small
\setlength{\tabcolsep}{3pt}     
\renewcommand{\arraystretch}{1.1}
\begin{tabularx}{\textwidth}{@{}%
  >{\RaggedRight\arraybackslash}p{0.18\textwidth}%
  >{\RaggedRight\arraybackslash}p{0.28\textwidth}%
  >{\RaggedRight\arraybackslash}X@{}}
\toprule
\textbf{Task} & \textbf{Template} & \textbf{Example} \\ \midrule
Emo6 & COTs \newline \textit{(Chain-of-Thought: Analyzes image elements like shape, color, texture, or contextual cues to support the given emotion distribution)} & The image depicts a house adorned with elaborate Halloween decorations at night, illuminated by red lights. Key objects include large, life-sized figurines of ghosts, ghouls, and other eerie figures, some of which appear to be holding candles. The decorations are meticulously arranged, creating a festive and somewhat eerie atmosphere. The shadows cast by the figures and the tree branches add depth to the scene, enhancing the sense of mystery and suspense. The overall emotional probability distribution suggests a mix of neutral and slightly negative emotions. The presence of the candles and the eerie figures might contribute to the feeling of unease, aligning with the neutral and slightly negative emotions inferred from the distribution. \\[2pt] \hline
LISA & target \newline \textit{(Unique per sample: Comprises the target object’s name and its descriptive attributes)} & The object in the picture that could be used to block the sun’s glare and protect the eyes\\[2pt] \hline
LISA & COTs \newline \textit{(Chain-of-Thought: Identifies the relationship between the target description and its location, forming evidence for selecting the bounding box)} & 1. The object in question is a piece of fabric because it is enclosed within the given rectangular coordinates.
2. It is evident from the color and texture that the fabric is a type of object to block the sun's glare and protect the eyes.
3. The object has the appearance of being worn or hung around the neck, which is characteristic of a typical piece of fabric.
4. The surrounding context suggests that it is a dog, as evidenced by the reflection in the glasses and the overall appearance which is consistent with a dog.
5. Therefore, the logical conclusion is that this region corresponds to the target object, which is a piece of fabric used to protect the eyes.
Hence, the object that corresponds to this region is a piece of fabric. \\ \bottomrule
\end{tabularx}
\caption{RISE-CoT Reasoning and Target Examples.The examples are drawn from the first of the three targets in Figure 1 and Figure 2, respectively.}
\label{tab:template_vars}
\end{table*}

\subsubsection{Annotation Reconstruction Prompts}

The Annotation Reconstruction step prompts the VLM to predict the annotation ($\hat{A}_i$) based solely on the CoT ($R_i$) and the image ($I_i$), verifying the CoT's sufficiency. The prompts are structured as follows:
\begin{itemize}
    \item \textbf{Classification (Emotion6):} 
    \textit{Analyze the provided image description and generate a probability distribution across emotion categories. Categories: ['anger', 'disgust', 'fear', 'joy', 'sadness', 'surprise', 'neutral']. Description: \textbf{\$CoTs\$}. Output the final answer in the following format: \texttt{<answer>\{'anger': prob\_0, 'disgust': prob\_1, 'fear': prob\_2, 'joy': prob\_3, 'sadness': prob\_4, 'surprise': prob\_5, 'neutral': prob\_6\}</answer>}. Here, \textnormal{prob\_0, prob\_1, \dots, prob\_6} represent the probabilities for each emotion category, and they should sum to 1. Do not provide any additional explanations or reasoning. Only return the result in the specified format.}

    \item \textbf{Detection (LISA):} 
    \textit{According to the following reasoning description of a region corresponding to \textbf{\$target\$} in the image: \textbf{\$CoTs\$}, output its specific bounding box in the format: \texttt{<answer>[x1, y1, x2, y2]</answer>}. Please strictly follow the format.}
\end{itemize}

These prompts ensure that the CoTs are visually grounded and logically consistent, while the reconstruction step verifies their sufficiency for accurate annotation prediction.

\subsection{RISE-R1}
The RISE-R1 stage designs prompts to initialize the reasoning process for both classification and detection tasks, emphasizing format compliance and reasoning.
\begin{itemize}
    \item \textbf{Classification (Emotion6):} \\
    \textit{Analyze the visual elements and interactions in the image to explain the emotional atmosphere. Then, generate a probability distribution across the following emotion categories: \{'anger', 'disgust', 'fear', 'joy', 'sadness', 'surprise', 'neutral'\}. Please output the final answer in the following format: \texttt{<think>\ldots</think><answer>\{'anger': prob\_0, 'disgust': prob\_1, 'fear': prob\_2, 'joy': prob\_3, 'sadness': prob\_4, 'surprise': prob\_5, 'neutral': prob\_6\}</answer>}. Here, prob\_0, prob\_1, \ldots, prob\_6 represent the probabilities for each emotion category, and they should sum to 1.}

    \item \textbf{Detection (LISA):} \\
    \textit{Given an image, identify the region corresponding to \{target\}. Provide a structured output with a reasoning explanation followed by the predicted bounding box coordinates. Format the output as: \texttt{<think>Analyze the shape, color, texture, and surrounding objects that help localize the \{target\} in the image.</think><answer>[x1, y1, x2, y2]</answer>}}
\end{itemize}

\section{Appendix B: Reward Distributions and Threshold Analysis}

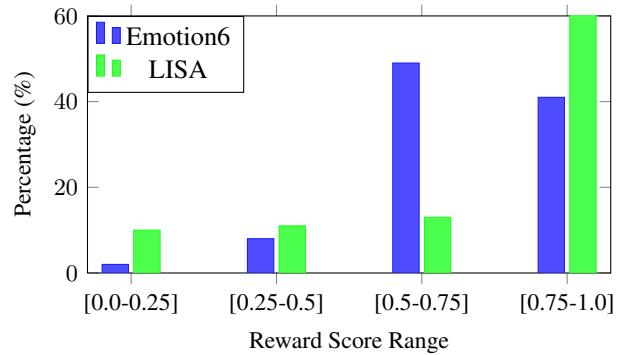
\begin{figure}[htbp]
    \centering
    \begin{tikzpicture}
        \begin{axis}[
            ybar,
            width=0.48\textwidth,
            height=5cm,
            bar width=0.35cm,
            symbolic x coords={[0.0-0.25], [0.25-0.5], [0.5-0.75], [0.75-1.0]},
            xtick=data,
            xlabel={Reward Score Range},
            ylabel={Percentage (\%)},
            ymin=0, ymax=60,
            legend style={at={(0.0,1.0)}, anchor=north west},
            axis line style={-},
            tick label style={font=\small},
            label style={font=\small},
            title style={font=\small}
        ]
        \addplot[fill=blue!70, draw=blue!90] coordinates {
            ([0.0-0.25], 2)
            ([0.25-0.5], 8)
            ([0.5-0.75], 49)
            ([0.75-1.0], 41)
        };
        \addplot[fill=green!70, draw=green!90] coordinates {
            ([0.0-0.25], 10)
            ([0.25-0.5], 11)
            ([0.5-0.75], 13)
            ([0.75-1.0], 66)
        };
        \legend{Emotion6, LISA}
        \end{axis}
    \end{tikzpicture}
    \captionsetup{style=ruled}
    \caption{Reward score distribution for Emotion6 and LISA.}
    \label{fig:reward_distribution}
\end{figure}

The RISE-CoT stage generates Chains of Thought (CoTs) (\(R^*\)), reconstructed annotations (\(\hat{A}\)), and reward scores (\(r \in [0.0, 1.0]\)) using the reward function \(\mathcal{R}(A_i, \hat{A}_i, R_i)\). Figure~\ref{fig:reward_distribution} shows the reward score distribution for Emotion6 and LISA, with 41\% (568/1,386) and 66\% (231/350) of samples having \(r \geq 0.75\), respectively, indicating that \(\tau = 0.75\) captures a significant proportion of high-quality CoTs across tasks of varying reasoning difficulty.

\begin{figure}[htbp]
    \centering
    \begin{tikzpicture}
        \begin{axis}[
            ybar,
            width=0.48\textwidth,
            height=5cm,
            bar width=0.35cm,
            symbolic x coords={[0.0-0.25], [0.25-0.5], [0.5-0.75], [0.75-1.0]},
            xtick=data,
            xlabel={Reward Score Range},
            ylabel={Percentage (\%)},
            ymin=0, ymax=60,
            legend style={at={(0.0,1.0)}, anchor=north west},
            axis line style={-},
            tick label style={font=\small},
            label style={font=\small},
            title style={font=\small}
        ]
        \addplot[fill=blue!70, draw=blue!90] coordinates {
            ([0.0-0.25], 25)
            ([0.25-0.5], 66)
            ([0.5-0.75], 6)
            ([0.75-1.0], 3)
        };
        \addplot[fill=green!70, draw=green!90] coordinates {
            ([0.0-0.25], 61)
            ([0.25-0.5], 21)
            ([0.5-0.75], 17)
            ([0.75-1.0], 1)
        };
        \legend{Emotion6, LISA}
        \end{axis}
    \end{tikzpicture}
    \captionsetup{style=ruled}
    \caption{Reward score distribution of \textbf{corrupted samples} for Emotion6 and LISA.}
    \label{fig:reward_distribution_noise}
\end{figure}
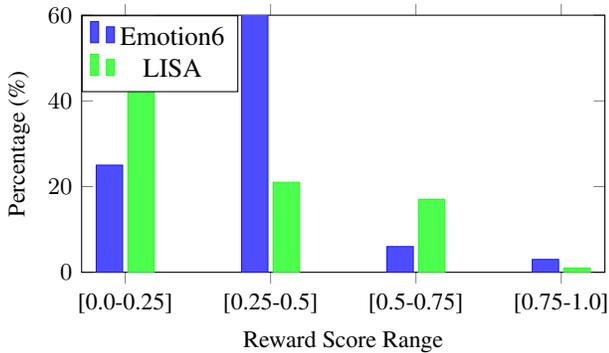

To test robustness, we corrupted 30\% of labels in each dataset. For Emotion6, we assigned a new random probability distribution, ensuring the highest-probability category differs from the original. For LISA, we selected a random bounding box region that does not overlap with the original. Figure~\ref{fig:reward_distribution_noise} shows the reward score distribution for corrupted samples, with over 95\% of corrupted samples having \(r < 0.75\), demonstrating that \(\tau = 0.75\) effectively filters noisy data.

\textbf{Analysis}: The choice of \(\tau = 0.75\) as the threshold for filtering high-quality CoTs is justified by its ability to retain a substantial portion of accurate samples (41\% for Emotion6, 66\% for LISA) while excluding over 95\% of noisy samples across both datasets. This balance ensures robust performance in diverse tasks. However, the effectiveness of \(\tau = 0.75\) is closely tied to the design of the reward function \(\mathcal{R}(A_i, \hat{A}_i, R_i)\), which evaluates the alignment between reconstructed and Ground Truth annotations. Changes to the reward function, such as altering the weighting of visual or contextual cues, could shift the optimal threshold, necessitating re-evaluation of \(\tau\) to maintain filtering efficacy.

\section{Appendix C: ImageNet-Sub Dataset Details}
For image classification tasks, we use a subset of ImageNet~\cite{deng2009imagenet}, termed ImageNet-Sub, comprising 20 classes: \textit{analog clock, backpack, ballpoint, Band Aid, barbell, barber chair, beer bottle, beer glass, binoculars, bolo tie, bookcase, bottlecap, brassiere, broom, bucket, buckle, candle, can opener, carton, and cellular telephone}. These classes were selected to ensure diversity in object types, covering everyday items (e.g., \textit{cellular telephone}, \textit{carton}), tools (e.g., \textit{can opener}, \textit{broom}), and specialized objects (e.g., \textit{barbell}, \textit{bolo tie}). Overall, these objects rely primarily on visual features such as shape, texture, and color for identification, requiring minimal complex reasoning. This selection balances visual distinctiveness and contextual variability, providing a representative subset for evaluating RISE’s image classification performance while maintaining computational efficiency. Each class contains 25 training and 10 testing images, totaling 500 training and 200 testing images, with one-hot probability distribution annotations.

\section{Appendix D: Results and Comparisons}
\subsection{RISE-CoT Intermediate Results}
\begin{figure}[htbp]
    \centering
    \includegraphics[width=\linewidth]{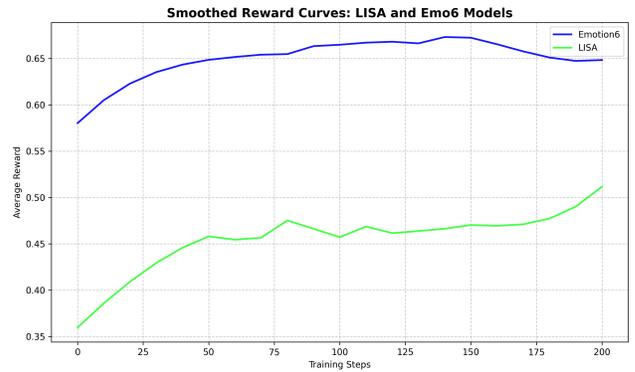}
    \captionsetup{style=ruled}
    \caption{Average reward curves for RISE-CoT training on Emotion6 and LISA, showing improved CoT quality over 200 steps.}
    \label{fig:cot_reward_curve}
\end{figure}

To illustrate the self-supervised learning process in RISE-CoT, we present the average reward $\mathcal{R}(A_i, \hat{A}_i, R_i)$ across training steps for Emotion6 and LISA datasets, as shown in Figure~\ref{fig:cot_reward_curve}. The curves demonstrate the improvement in CoT quality, with rewards converging to higher values as training progresses, reflecting the optimization of visually grounded and logically consistent CoTs.

To further demonstrate the improvement in CoT quality, we present the CoT evolution for one sample from each dataset at training steps 0, 100, and 200, along with their corresponding reward scores, as shown in Figure~\ref{fig:emo6_train} and Figure~\ref{fig:lisa_train}.

% \begin{figure}[htb]
%     \centering
%     % Placeholder for your figure showing CoT evolution
%     \captionsetup{style=ruled}
%     \caption{CoT evolution for sample images in Emotion6 and LISA during RISE-CoT training, showing CoT descriptions and reward scores at steps 0, 100, and 200.}
%     \label{fig:cot_evolution}
% \end{figure}

\textit{Discussion}: The CoT evolution illustrates how RISE-CoT refines vague initial descriptions into detailed, visually grounded reasoning chains, leading to higher rewards due to improved reconstruction accuracy. The reward increase reflects the model’s ability to generate CoTs that better align with the Ground Truth annotations.

\begin{figure*}[htb]
    \centering
    \includegraphics[width=\textwidth]{LaTeX/IMD_COCO_COT.drawio.png} 
    \captionsetup{style=ruled}
    \caption{Extended "think-answer" comparisons for RISE-R1 on ImageNet-Sub, and COCO-Sub, showing outputs of RISE, Base-Model, SFT, Visual-RFT, and GPT-4o.}
    \label{fig:r1_results}
\end{figure*}

\subsubsection{Failure Case Analysis}
To highlight the limitations of RISE-CoT, we analyze samples with low reward scores, indicating challenges in generating high-quality CoTs. Figure~\ref{fig:emo6_cot_failure} and Figure~\ref{fig:lisa_cot_failure} presents two failure cases from Emotion6 and LISA, including the original images, Ground Truth annotations, reconstructed annotations, and corresponding reward values. These cases typically involve images with low clarity (e.g., blurry visuals or ambiguous objects), which hinder the model’s ability to extract distinct visual cues, resulting in less accurate CoTs and lower rewards. This analysis underscores RISE-CoT’s dependency on clear visual information for effective reasoning and reconstruction, consistent with the limitations discussed in the paper.

\subsection{RISE-R1 Training Process and Results}

\begin{figure}[H]
    \centering
    \begin{tikzpicture}
        \begin{axis}[
            xlabel={Training Steps},
            ylabel={Loss (SFT) / Reward (RFT)},
            grid=major,
            % 将图例放置在正上方
            legend style={
                at={(0.5,1.05)}, % 水平居中，垂直位置在图形上方5%
                anchor=south,    % 锚点在底部中点
                legend columns=2, % 4个条目分成4列
                /tikz/every even column/.append style={column sep=5pt},
                font=\footnotesize
            },
            width=0.8\linewidth,
            height=5cm,
            tick label style={font=\small},
            label style={font=\small},
            title style={font=\small}
        ]
       
        \addlegendentry{Emotion6 (SFT Loss)}
        \addplot[green, thick,dotted] coordinates {
    (10, 1.2298) (20, 1.1565) (30, 1.0295) (40, 0.9738) (50, 0.9267)
    (60, 0.9037) (70, 0.8454) (80, 0.7989) (90, 0.7564) (100, 0.7264)
    (110, 0.6536) (120, 0.6207) (130, 0.5491) (140, 0.5369) (150, 0.4865)
    (160, 0.4648) (170, 0.4522) (180, 0.4372) (190, 0.4331) (200, 0.4294)
};
        \addlegendentry{LISA (SFT Loss)}
        % RFT Reward (Emotion6 and LISA)
        \addplot[blue, thick,dotted] coordinates
        {
    (10, 1.7034) (20, 1.4891) (30, 1.3207) (40, 1.2230) (50, 1.1591)
    (60, 1.0898) (70, 1.0411) (80, 0.9434) (90, 0.8568) (100, 0.7552)
    (110, 0.6484) (120, 0.5660) (130, 0.4785) (140, 0.4199) (150, 0.3592)
    (160, 0.3317) (170, 0.2982) (180, 0.2900) (190, 0.2842) (200, 0.2794)
};
        \addlegendentry{Emotion6 (RFT Reward)}
        \addplot[green, thick] coordinates {
            (200, 0.12) (225, 0.24) (250, 0.36) (275, 0.55) (400, 0.82)
        };
        \addlegendentry{LISA (RFT Reward)}
        \addplot[blue, thick] coordinates {
            (200, 0.21) (225, 0.35) (250, 0.44) (275, 0.51) (400, 0.77)
        };
        \draw[dashed] (axis cs:200,0) -- (axis cs:200,2.5);
        \node at (axis cs:50,2.3) {SFT};
        \node at (axis cs:150,2.3) {RFT};
        \end{axis}
    \end{tikzpicture}
    \captionsetup{style=ruled}
    \caption{Training curves for RISE-R1 on Emotion6 and LISA, showing SFT loss (0–200 steps, dashed lines) and RFT reward (200–400 steps, solid lines).}
    \label{fig:r1_training_curves}
\end{figure}

The RISE-R1 stage consists of Supervised Fine-Tuning (SFT) on $\mathcal{D}_\text{RISE}^\text{high}$ (first 200 steps) followed by Reinforcement Fine-Tuning (RFT) on $\mathcal{D}$ (next 200 steps). Figure~\ref{fig:r1_training_curves} shows the training dynamics: the SFT phase with the average loss (e.g., cross-entropy loss) for Emotion6 and LISA, and the RFT phase with the average reward $\mathcal{R}_\text{R1}(A_i, \hat{A}_i^{R1}, R_i^{R1})$. The decreasing loss in SFT reflects improved annotation accuracy using high-quality CoTs, while the increasing reward in RFT indicates enhanced CoT interpretability and alignment with Ground Truth annotations.

To showcase RISE-R1’s final performance, we extend the qualitative comparisons in Figure 3 (main paper) with additional "think-answer" outputs for ImageNet-Sub, and COCO-Sub, comparing RISE with Base-Model, SFT, Visual-RFT, and GPT-4o, as shown in Figure~\ref{fig:r1_results}. These results highlight RISE-R1’s ability to generate detailed, visually grounded CoTs, leading to accurate annotations across diverse tasks.

% \begin{figure}[htb]
%     \centering
%     % Placeholder for your figure showing Think-Answer results
%     \includegraphics[width=\columnwidth]{LaTeX/IMD_COCO_COT.drawio.png}
%     \captionsetup{style=ruled}
%     \caption{Extended "think-answer" comparisons for RISE-R1 on ImageNet-Sub, and COCO-Sub, showing outputs of RISE, Base-Model, SFT, Visual-RFT, and GPT-4o.}
%     \label{fig:r1_results}
% \end{figure}

% To illustrate RISE-R1’s limitations, we present failure cases with low annotation accuracy, as shown in Figure~\ref{fig:r1_failure}. These cases, typically involving blurry images or complex scenes, result in less precise CoTs and annotations, underscoring RISE-R1’s dependency on clear visual cues, consistent with the limitations discussed in Section 5.

% \begin{figure}[htb]
%     \centering
%     % Placeholder for your figure showing RISE-R1 failure cases
%     \captionsetup{style=ruled}
%     \caption{Failure cases in RISE-R1 for Emotion6 and LISA, showing original image descriptions, Ground Truth annotations, RISE-R1’s "think-answer" outputs, and low accuracy due to unclear visual cues.}
%     \label{fig:r1_failure}
% \end{figure}

\begin{figure*}[hbtp]
    \centering
    \includegraphics[width=0.95\linewidth]{LaTeX/emo6_cot.pdf}
    \captionsetup{style=ruled}
    \caption{ CoT evolution for sample images in Emotion6 during RISE-CoT  training, showing CoT descriptions and reward scores at steps 0, 100, and 200.}
    \label{fig:emo6_train}
\end{figure*}

\begin{figure*}[hbtp]
    \centering
    \includegraphics[width=0.95\linewidth]{LaTeX/LISA_cot.pdf}
    \captionsetup{style=ruled}
    \caption{ CoT evolution for sample images in LISA during RISE-CoT  training, showing CoT descriptions and reward scores at steps 0, 100, and 200.}
    \label{fig:lisa_train}
\end{figure*}

\begin{figure*}[htb]
    \centering
    \includegraphics[width=0.95\linewidth]{LaTeX/EMO6_cot_false.pdf}
    % Placeholder for your figure showing failure cases
    \captionsetup{style=ruled}
    \caption{Failure cases in RISE-CoT for Emotion6, showing original image descriptions, Ground Truth annotations, reconstructed annotations, and low reward scores due to unclear visual cues.}
    \label{fig:emo6_cot_failure}
\end{figure*}

\begin{figure*}[htb]
    \centering
    \includegraphics[width=0.95\linewidth]{LaTeX/LISA_cot_false.pdf}
    % Placeholder for your figure showing failure cases
    \captionsetup{style=ruled}
    \caption{Failure cases in RISE-CoT for LISA, showing original image descriptions, Ground Truth annotations, reconstructed annotations, and low reward scores due to unclear visual cues.}
    \label{fig:lisa_cot_failure}
\end{figure*}

\bibliography{aaai2026}